\documentclass[a4paper,fleqn]{cas-sc}

\usepackage[authoryear,longnamesfirst]{natbib}

\graphicspath{{figs/}} 
\usepackage{amsmath} 
\usepackage{physics} 
\usepackage[ruled,vlined]{algorithm2e} 
\usepackage{graphicx}
\usepackage{epstopdf,epsfig}
\usepackage{hyperref}
\usepackage{amssymb}
\newtheorem{theorem}{Theorem} 
\usepackage{booktabs}
\usepackage{multirow}
\usepackage{siunitx}
\usepackage{tabularx}
\usepackage[symbols, acronym, nopostdot, style=super, nonumberlist, toc]{glossaries}
\newglossaryentry{G}{
    sort={G},
    name={\ensuremath{G}},
    description={Gravitational acceleration (Chapter 2) \glspar Graph (Chapter 3)},
    type=symbols
}
\newglossaryentry{E}{
    sort={E},
    name={\ensuremath{E}},
    description={Hencky elastic strain (Chapter 2) \glspar Set of graph edge features or embeddings (Chapter 3)},
    type=symbols
}
\newglossaryentry{V}{
    sort={V},
    name={\ensuremath{V}},
    description={Particle volume (Chapter 2) \glspar Set of graph node features or embeddings (Chapter 3)},
    type=symbols
}
\newglossaryentry{u}{
    sort={u},
    name={\ensuremath{u}},
    description={Graph global feature},
    type=symbols
}
\newglossaryentry{e}{
    sort={e},
    name={\ensuremath{e}},
    description={Graph edge features},
    type=symbols
}
\newglossaryentry{v}{
    sort={v},
    name={\ensuremath{v}},
    description={Velocity (Chapter 2) \glspar Graph node features (Chapter 3)},
    type=symbols
}
\newglossaryentry{ri}{
    sort={r},
    name={\ensuremath{r_i}},
    description={Index of graph receiver node of edge $i$},
    type=symbols
}
\newglossaryentry{si}{
    sort={s},
    name={\ensuremath{s_i}},
    description={Index of graph sender node of edge $i$},
    type=symbols
}
\newglossaryentry{Ne}{
    sort={Ne},
    name={\ensuremath{N^e}},
    description={Number of graph edges},
    type=symbols
}
\newglossaryentry{Nv}{
    sort={Nv},
    name={\ensuremath{N^v}},
    description={Number of graph nodes},
    type=symbols
}
\newglossaryentry{hu}{
    sort={hu},
    name={\ensuremath{h_u}},
    description={Graph global embeddings},
    type=symbols
}
\newglossaryentry{he}{
    sort={he},
    name={\ensuremath{h_e}},
    description={Graph edge embeddings},
    type=symbols
}
\newglossaryentry{hv}{
    sort={hv},
    name={\ensuremath{h_v}},
    description={Graph node embeddings},
    type=symbols
}
\newglossaryentry{N}{
    sort={N},
    name={\ensuremath{N}},
    description={Quadratic B-splines (Chapter 2) \glspar Graph node neighborhood (Chapter 3)},
    type=symbols
}
\newglossaryentry{Nk}{
    sort={Nk},
    name={\ensuremath{N^k}},
    description={Number of message-passing steps},
    type=symbols
}
\newglossaryentry{S}{
    sort={S},
    name={\ensuremath{S}},
    description={Material shear modulus (Chapter 2) \glspar Flow states (Chapter 3)},
    type=symbols
}
\newglossaryentry{Sp}{
    sort={Sp},
    name={\ensuremath{S^{\prime}}},
    description={Reduced flow states},
    type=symbols
}
\newglossaryentry{Sbar}{
    sort={Sbar},
    name={\ensuremath{\bar S}},
    description={Mean of flow states},
    type=symbols
}
\newglossaryentry{B}{
    sort={B},
    name={\ensuremath{B}},
    description={Material bulk modulus (Chapter 2) \glspar Mean-subtracted data (Chapter 3)},
    type=symbols
}
\newglossaryentry{C}{
    sort={C},
    name={\ensuremath{C}},
    description={APIC mode coefficient (Chapter 2) \glspar Covariance (Chapter 3)},
    type=symbols
}
\newglossaryentry{R}{
    sort={R},
    name={\ensuremath{R}},
    description={Rotation part of deformation gradient (Chapter 2) \glspar Rigid states (Chapter 3)},
    type=symbols
}
\newglossaryentry{m}{
    sort={m},
    name={\ensuremath{m}},
    description={Particle mass (Chapter 2) \glspar Number of observations in data (Chapter 3)},
    type=symbols
}
\newglossaryentry{n}{
    sort={n},
    name={\ensuremath{n}},
    description={Number of dimensions in data},
    type=symbols
}
\newglossaryentry{r}{
    sort={r},
    name={\ensuremath{r}},
    description={Number of subspace modes},
    type=symbols
}
\newglossaryentry{U}{
    sort={U},
    name={\ensuremath{U}},
    description={Stretch part of deformation gradient (Chapter 2) \glspar PCA transformation (Chapter 3)},
    type=symbols
}
\newglossaryentry{X}{
    sort={X},
    name={\ensuremath{X}},
    description={Position in undeformed state, spatial coordinate (Chapter 2) \glspar MPM Data (Chapter 3)},
    type=symbols
}
\newglossaryentry{F}{
    sort={F},
    name={\ensuremath{F}},
    description={Deformation gradient (Chapter 2) \glspar Interaction force applied to COM of rigid body (Chapter 3)},
    type=symbols
}
\newglossaryentry{Z}{
    sort={Z},
    name={\ensuremath{Z}},
    description={Reduced flow and rigid states},
    type=symbols
}
\newglossaryentry{c}{
    sort={c},
    name={\ensuremath{c}},
    description={Number of examples in each dataset},
    type=symbols
}
\newglossaryentry{d}{
    sort={d},
    name={\ensuremath{d}},
    description={Grain diameter (Chapter 2) \glspar Number of history timesteps (Chapter 3)},
    type=symbols
}
\newglossaryentry{lambda}{
    sort={1lambda},
    name={\ensuremath{\lambda}},
    description={Eigenvalue},
    type=symbols
}
\newglossaryentry{o}{
    sort={o},
    name={\ensuremath{o}},
    description={GNS Decoder output},
    type=symbols
}
\newglossaryentry{a}{
    sort={a},
    name={\ensuremath{a}},
    description={Node/subspace particle acceleration},
    type=symbols
}
\newglossaryentry{z}{
    sort={z},
    name={\ensuremath{z}},
    description={Node/subspace particle position},
    type=symbols
}
\newglossaryentry{L}{
    sort={L},
    name={\ensuremath{L}},
    description={Velocity spatial gradient (Chapter 2) \glspar Loss function (Chapter 3)},
    type=symbols
}
\newglossaryentry{theta}{
    sort={1theta},
    name={\ensuremath{\theta}},
    description={GNS model parameters},
    type=symbols
}

\newglossaryentry{f}{
    sort={f},
    name={\ensuremath{f}},
    description={Particle internal force},
    type=symbols
}
\newglossaryentry{T}{
    sort={T},
    name={\ensuremath{T}},
    description={Cauchy stress},
    type=symbols
}
\newglossaryentry{I}{
    sort={I},
    name={\ensuremath{I}},
    description={Inertial number},
    type=symbols
}
\newglossaryentry{mu}{
    sort={1mu},
    name={\ensuremath{\mu}},
    description={Friction coefficient},
    type=symbols
}
\newglossaryentry{mu2}{
    sort={1mu2},
    name={\ensuremath{ \mu_2 }},
    description={Maximum friction coefficient (saturation value)},
    type=symbols
}
\newglossaryentry{mus}{
    sort={1mus},
    name={\ensuremath{ \mu_\text{s} }},
    description={Minimum (static) friction coefficient},
    type=symbols
}
\newglossaryentry{gammadotp}{
    sort={1gammad},
    name={\ensuremath{ \dot{\gamma}^\text{p} }},
    description={Equivalent plastic shear strain rate},
    type=symbols
}
\newglossaryentry{gamma}{
    sort={1gamma},
    name={\ensuremath{ \gamma }},
    description={Shear strain},
    type=symbols
}
\newglossaryentry{rhos}{
    sort={1rhos},
    name={\ensuremath{ \rho_\text{s} }},
    description={Grain density},
    type=symbols
}
\newglossaryentry{p}{
    sort={p},
    name={\ensuremath{ p }},
    description={Mean normal stress},
    type=symbols
}
\newglossaryentry{phi}{
    sort={1phi},
    name={\ensuremath{ \phi }},
    description={Volume fraction},
    type=symbols
}
\newglossaryentry{g}{
    sort={g},
    name={\ensuremath{ g }},
    description={Granular fluidity},
    type=symbols
}
\newglossaryentry{A}{
    sort={A},
    name={\ensuremath{ A }},
    description={Nonlocal amplitude},
    type=symbols
}
\newglossaryentry{tau}{
    sort={1tau},
    name={\ensuremath{ \tau }},
    description={Shear stress},
    type=symbols
}
\newglossaryentry{rho}{
    sort={1rho},
    name={\ensuremath{ \rho }},
    description={Density},
    type=symbols
}
\newglossaryentry{q}{
    sort={q},
    name={\ensuremath{ q }},
    description={Test function},
    type=symbols
}
\newglossaryentry{Phi}{
    sort={1Phi},
    name={\ensuremath{ \Phi }},
    description={Shape function},
    type=symbols
}
\newglossaryentry{x}{
    sort={x},
    name={\ensuremath{ x }},
    description={Position in deformed state, material coordinate},
    type=symbols
}
\newglossaryentry{P}{
    sort={P},
    name={\ensuremath{ P }},
    description={Polynomial basis},
    type=symbols
}
\newglossaryentry{kappa}{
    sort={1kappa},
    name={\ensuremath{ \kappa }},
    description={Localized weighting function},
    type=symbols
}
\newglossaryentry{M}{
    sort={M},
    name={\ensuremath{ M }},
    description={Mandel stress},
    type=symbols
}
\newglossaryentry{D}{
    sort={D},
    name={\ensuremath{ D }},
    description={Stretch part of velocity spatial gradient},
    type=symbols
}

\begin{document}
\let\WriteBookmarks\relax
\def\floatpagepagefraction{1}
\def\textpagefraction{.001}
\shorttitle{Subspace Graph Physics}
\shortauthors{A. Haeri et~al.}

\title [mode = title]{Subspace Graph Physics: Real-Time Rigid Body-Driven Granular Flow Simulation}                      
\author[1]{Amin Haeri}[orcid=0000-0002-1217-656X]
\ead{amin.haeri@concordia.ca}

\author[1]{Krzysztof Skonieczny}[orcid=0000-0002-6540-3922]
\ead{krzysztof.skonieczny@concordia.ca}
\cormark[1]

\address[1]{Department of Electrical and Computer Engineering, Concorida University, Canada}

\cortext[cor1]{Corresponding author}

\begin{abstract}
    An important challenge in robotics is understanding the interactions between robots and deformable terrains that consist of granular material. Granular flows and their interactions with rigid bodies still pose several open questions. A promising direction for accurate, yet efficient, modeling is using continuum methods. Also, a new direction for real-time physics modeling is the use of deep learning. This research advances machine learning methods for modeling rigid body-driven granular flows, for application to terrestrial industrial machines as well as space robotics (where the effect of gravity is an important factor). In particular, this research considers the development of a subspace machine learning simulation approach. To generate training datasets, we utilize our high-fidelity continuum method, material point method (MPM). Principal component analysis (PCA) is used to reduce the dimensionality of data. We show that the first few principal components of our high-dimensional data keep almost the entire variance in data. A graph network simulator (GNS) is trained to learn the underlying subspace dynamics. The learned GNS is then able to predict particle positions and interaction forces with good accuracy. More importantly, PCA significantly enhances the time and memory efficiency of GNS in both training and rollout. This enables GNS to be trained using a single desktop GPU with moderate VRAM. This also makes the GNS real-time on large-scale 3D physics configurations (700x faster than our continuum method). 
\end{abstract}


\maketitle

\section{Introduction}
    
    Granular flows and their interactions with rigid bodies still pose several open questions. Their modeling is complex as they can experience various solid-like, fluid-like and gas-like deformations in time. Engineering applications of machine-terrain interactions include earthwork using industrial excavators, bulldozers, etc. as well as agricultural vehicles. Accurately modeling how these machines interact with granular materials can lead to the development of better training simulators and to the automation of tasks. Robot-terrain interactions are also very important in space exploration, where rovers drive over granular regolith in the reduced-gravity environments of Mars and the Moon. The entrapment of the Mars Exploration Rover Spirit in soft regolith and the tears and punctures in the Mars Science Laboratory Curiosity rover’s wheels demonstrate some of the current challenges of such reduced-gravity granular terrains. Therefore, a real-time and accurate simulation method is essential for robot mobility control, training operators, or for eventually automating aspects of these operations in the construction and space industries.
    
    Current robot-terrain interaction models have very high computational complexity, depend on many difficult to measure parameters, and/or have insufficient predictive power. In terms of high accuracy, one current direction of research is the discrete element method (DEM), which demonstrates promise in modeling granular flows. But it is so computationally intensive as to be infeasible for online applications \citep{comp13}, and for large physical domains can be untenably expensive even in offline applications \citep{DunDemExpK}. On the other end of the complexity spectrum, several researchers today highlight the insufficient predictive power of classical terramechanics models \citep{comp14,comp15}, and their limitations to specific flow geometries \citep{Hae20isarc}, though they can be real-time. Even modern continuum methods such as material point method (MPM), that simultaneously achieve accuracy and relative efficiency, still may not be real-time for 3D configurations with existing computer processors, especially in hardware-constrained applications like planetary rovers \cite{Hae21eas}.
    
    However, data-driven methods have begun to be applied to similar (especially fluid) simulations. Some researchers \citep{pcamg, subs1, subs2, subs3, subs4} are applying dimensionality reduction methods, such as principal component analysis (PCA) \citep{pcaref2}, to get a lower-dimensional representation of the physics data. Others \citep{gns, mitgraph} are using graph neural networks (GNNs) \citep{gnn0, gnn, gnn1, gnn2}, such as graph networks (GNs) \citep{gn}, to learn the underlying physical dynamics more accurately. Metrics of success for these methods have to date focused on qualitative visual appearance. In the following, three important aspects of such approaches are reviewed: data, reduced data, and learning.
       
    \textit{Data:}
        Since machine learning physics simulation methods require hours of data at various initial and boundary conditions, numerical methods will be needed to complement and may even outweigh the use of experiments. To generate fluid simulations as training data, several lines of work have utilized smoothed particle hydrodynamics (SPH \citep{Mon92sph}) \citep{gns}, fluid implicit particle (FLIP \citep{Bri15flip}) \citep{Kim19cnn,Um18ff}, finite volume method (FVM) \citep{Thu20ae}, or finite difference method (FDM) \citep{Ozb19pde}. Also, some general methods such as position based dynamics (PBD \citep{Mac14flex}) \citep{pcamg,mitgraph}, and projective dynamics (PD \citep{Bou14pd}) \citep{Nar16pduse,Wei16pduse} can be used for different types of material simulation (e.g. sand, snow, jelly). However, the continuum hybrid (i.e. particle and mesh-based) force-based material point method (MPM) with an accurate material model (e.g. nonlocal granular fluidity, NGF \citep{UnsNonK}), is as one of the most accurate and efficient methods for our purpose of granular flow modeling. MPM \citep{mpmHu} has also been recently used to produce sand simulations for the purpose of training a simulator \citep{gns}.

    \textit{Reduced Data:}
        In physics simulations, model reduction methods are utilized to capture the effective degrees of freedom of a physics system \citep{subs1,subs2,subs3,subs4}. This is with the hope to reduce both processing time and memory space usage corresponding to the system. Some work has also applied subspace methods to the equations of motion in FEM solvers \citep{subsfem1, subsfem2}. However, it is difficult to handle collisions in such approaches, and further modifications \citep{subscol} significantly reduce its efficiency. Recent literature \citep{pcamg} has shown that principal component analysis (PCA) could be an effective choice to capture the primary modes of elastic object deformations. 

    \textit{Learning:}
        Machine learning methods have been recently shown as real-time alternatives to traditional numerical methods to learn from data. One crucial challenge here is to properly learn the underlying dynamics principles in the physical systems automatically. Some methods have been applied to differential equations such as Poisson equation \citep{Ozb19pde}.
        Several recent methods have also been utilized for fluid simulations, including: regression forest \citep{Lad17regFor,Lad15regFor}, multi layer perceptron (MLP) \citep{Wie19ff,Um18ff}, autoencoder (AE) \citep{Thu20ae,Wan19ae,Mor18ae} convolutional neural network (CNN) \citep{Kim19cnn,Tom17cnn}, continuous convolution (CConv) \citep{Umm20cc}, generative adversarial network (GAN) \citep{Xie18gan}, and loss function-based method \citep{Pra20lf}. These methods are useful for a specific state of materials (gas-like, solid-like, or specifically fluid-like). They would likely need modifications to the architecture itself (not to mention additional data as well, of course) to make them work in the other states. However, graph networks (GN) \citep{gn}, a type of graph neural network \citep{gnn}, have been shown to be capable of simulating different states of materials while being simple to implement. Dynamic interaction network \citep{mitgraph}, as a GN variant, has been proposed for general simulations including elastic and rigid objects. But, it also requires additional computations and data to simulate elasto-plastic materials. In fact, the current position and velocity are no longer sufficient as inputs; additionally, the resting position and an extra network type (hierarchical modeling) are required. Recent state-of-the-art research has developed graph network simulator (GNS) \citep{gns}, which has outperformed some other recent methods \citep{mitgraph,Umm20cc} in terms of accuracy and ease of implementation.
    
    This research achieves accurate real-time engineering simulations of large-scale rigid body-driven granular flows. The metrics for accuracy include the positions of granular particles as well as the reaction forces on the rigid body. This is accomplished by integrating a high-accuracy machine learning approach, graph network simulator (GNS), with principal component analysis (PCA) into a subspace graph network simulator. GNS was customized to deal with subspace physics, to use a single fully-connected message passing step, and to predict interaction forces.  The system is trainable via a single desktop GPU with moderate VRAM. There are two distinct contributions here: (1) extensive training data generated via an accurate MPM verified by experimental data \citep{Hae20isarc,Hae21aero}, and (2) significantly reducing the training and rollout runtime and memory usage of GNs by learning the subspace dynamics (i.e. subspace learning \citep{SINDy}) using the reduced data obtained by applying PCA to high-dimensional physics data.
    
    This paper is organized in the following manner: we will first provide a background on graph neural networks (GNN). Then, it will develop various modules in the training phase of the subspace graph network simulator (GNS). The paper will end by providing quantitative and qualitative rollout results compared with the ground truth MPM results. The source code of the developed algorithm is available on GitHub as referred in the context.

\section{Graph Neural Networks} \label{subsec:GNN}
    This section provides a background on graph neural networks (GNNs). We introduce graph machine learning, and explain the GNNs (under the field of graph machine learning) and their necessity. Furthermore, we discuss the expressive power of GNNs and introduce graph network as generalized GNNs. We also explain one limitation of GNNs and why a non-learning model reduction method can help.
    
    \begin{figure}[!t]
        \centering
        \includegraphics[width=0.5\textwidth]{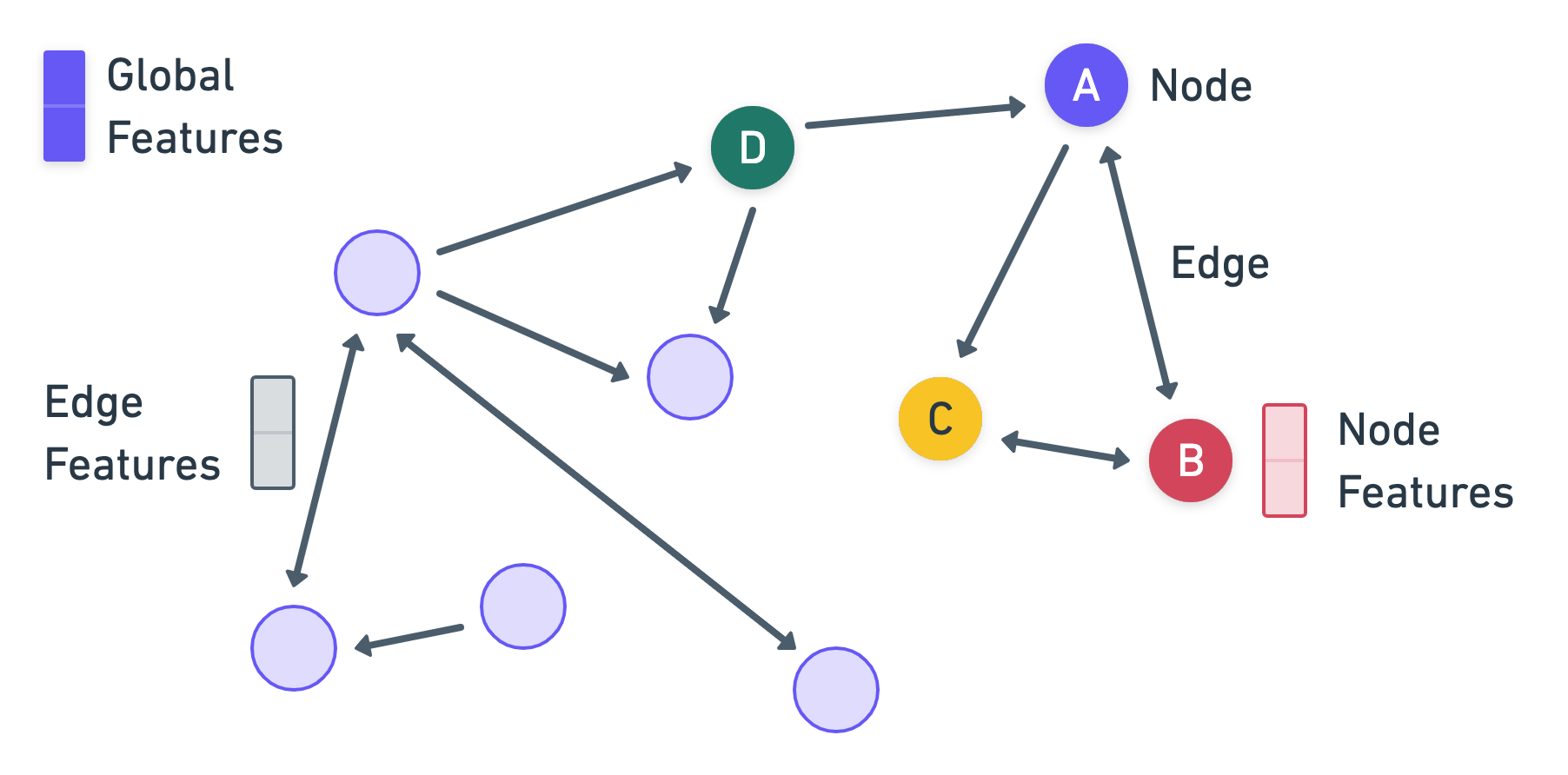}
        \includegraphics[width=0.4\textwidth]{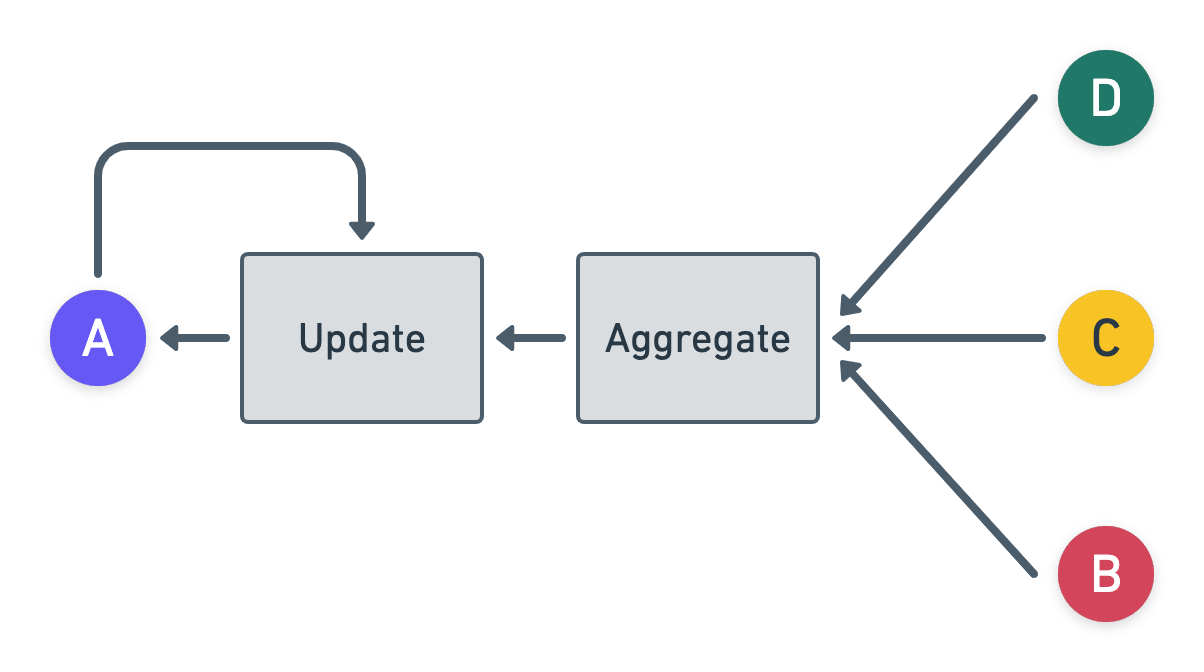}
        \caption{Left: A sample graph with nodes, directed edges, and features. Right: Illustration of one-step computational graph of node A in graph shown in Figure \ref{fig:graph}.}
        \label{fig:graph}
    \end{figure}
    
    Graphs are general data structures for describing complex systems. Graphs are composed of entities (also known as nodes) and pairwise interactions (edges). In many cases, graph objects can have attribute (feature) information differentiating them from each other. A sample graph is shown in Figure \ref{fig:graph}(Left). Machine learning with graphs enables us to do different node-, edge-, and graph-level tasks on graph data. These include, but are not limited to, node classification \citep{alphafold2}, relation prediction \citep{Ying_2018, bty294}, graph generation \citep{Konaklieva14Mol, StokesS20}, and graph evolution \citep{gns}. Traditional machine learning pipelines require \textit{hand-engineered} features extracted based on manually computed graph statistics and kernels (e.g. node degree, neighborhood overlap). These hand-engineered features are non-adaptive through a learning process, and time-consuming to process \citep{gnnbook}. Graph representation learning is an alternative approach to learning over graphs. In this approach, the aim is to automatically learn the features from encoded graph structural information.
    
    
    Graph neural networks (GNNs), in their basic form, are the generalization of convolutional neural network (CNN) beyond structured grid and sequence data to non-Euclidean data \citep{Bruna2014SpectralNA}. In fact, GNNs can handle data with arbitrary size and complex topological structure, varying node ordering, and often dynamic features. In a deep \textit{message-passing} framework, GNNs determine a node's \textit{computational graph} and propagate information through the graph. Below, we summarize how to propagate information across a graph to compute node features using a neural network.

    \subsection{Message-Passing Framework}
        In this framework, a graph is defined as $\gls{G}= (\gls{E}, \gls{V}, \gls{u})$. The \gls{u} is a global feature. The $\gls{V}= \{\gls{v}_i\}_{i=1:N^v}$ is a set of nodes where $\gls{v}_i$ are node features. The $\gls{E}= \{(\gls{e}_i, \gls{ri}, \gls{si})\}_{i=1:N^e}$ is a set of edges where $\gls{e}_i$ are edge features, $\gls{ri}$ is the index of receiver node, and $\gls{si}$ is the index of sender node. Also, \gls{Ne} and \gls{Nv} are the number of edges and nodes, respectively. Here the global, node and edge feature embeddings are $\gls{hu}= f_u(\gls{u})$, ${\gls{hv}}_i= f_v(\gls{v}_i)$ and ${\gls{he}}_i= f_e(\gls{e}_i)$, respectively, where $f(.)$'s are arbitrary functions. Note that the global and edge embeddings are only used in the graph network as a generalized message-passing framework.
    
        Every node has its own computational graph based on its local neighbourhood. The computational graph forms a tree structure by unfolding the neighborhood around the target node \citep{gnnbook}. It can have arbitrary depth (i.e. message-passing steps). A one-step computational graph for a single node is shown in Figure \ref{fig:graph}(Right). The message-passing framework is defined upon the computational graph. A basic framework consists of arbitrary differentiable node Update and Aggregate functions as follows
        \begin{equation}
            {\gls{hv}}_i^{(k+1)} = \text{Update}^{(k)} \left( {\gls{hv}_i}^{(k)}, \text{message}_{\gls{N}(\gls{v}_i)}^{(k)} \right)
            \label{equ:basic-mess-pass1}
        \end{equation}
        for 
        \begin{equation}
            \text{message}_{\gls{N}(\gls{v}_i)}^{(k)} = \text{Aggregate}^{(k)} \left(\{ {\gls{hv}}_j^{(k)} | \forall \gls{v}_j \in \gls{N}(\gls{v}_i)\} \right)
            \label{equ:basic-mess-pass2}
        \end{equation}
        where $\text{message}_{\gls{N}(\gls{v}_i)}^{(k)}$ is the message that is aggregated from node $\gls{v}_i$’s graph neighborhood, $\gls{N}(\gls{v}_i)$, at message-passing step $k$. Also, the Update function combines the aggregated message with the previous target node embeddings.
        
        The Update and Aggregate functions are applied per node. The Update function can be defined via a neural network (with nonlinear activation(s) to add expressiveness) in the GNN message-passing framework. There are different ways to aggregate the features, such as mean pooling in graph convolutional network (GCN) \citep{gcn}, and max pooling in GraphSage \citep{gsage}. Also, since the Aggregate function takes a set as input, the framework is permutation equivariant by design. This means that the Aggregate function output is permuted in a consistent way in response to permutations of the input \citep{gnnbook}. The expressiveness power of the Update function and the choice of the Aggregate function will be discussed later in the current section.
            
        In this framework GNN is inductive: (1) the parameters are shared, (2) the number of parameters is sublinear within the size of graph (e.g. node features size, $|\gls{V}|$), and (3) it can be generalized to unseen nodes. Specifically, GNNs are a certain class of general neural architectures. It means that graph structure is the input to the neural architecture (instead of being part of it), and the model parameters are shared to respect the invariance properties of the input graph \citep{pmlr-v119-you20b}. GNN examples include GCN, GraphSage, and Graph Attention Network (GAT) \citep{gat} with different message-passing architectures.

    \subsection{Expressiveness (Theory)}
        It is desirable for GNNs to have a high level of \textit{expressiveness}. The expressive power of GNNs is specified as the ability to distinguish different computational graphs (i.e. graph structures). \textit{Isomorphic} nodes are defined as the nodes that have the same computational graphs (i.e. same features and neighbourhood structure). GNNs are unable to distinguish two isomorphic nodes since they consider node features, not node IDs/indexes. Also, an \textit{injective} function maps every unique input to a different output. The most expressive GNN maps every computational graph into different node embeddings, injectively.
        
        Hence, the Aggregate function should be injective. That is, the expressiveness of GNNs can be characterized by the expressiveness of the Aggregate function over multisets (i.e. sets with repeating elements). For instance, GCN and GraphSage are not maximally powerful GNNs as they use non-injective mean and max pooling functions for aggregation, respectively. On the other hand, Theorem \ref{theo:gin} recently proposed by \cite{gin} indicates that a sum pooling operation is an injective function over multisets.
        \begin{theorem}
            Assume $\mathcal{X}$ is countable. There exists a function $f : \mathcal{X} \to \mathbb{R} ^n$ so that $h(X) = \sum_{x \in X} f(x)$ is unique for each multiset $X \subset \mathcal{X}$ of bounded size. Moreover, any multiset function $g$ can be decomposed as $g(X) = \Phi \left( \sum_{x \in X} f(x) \right)$ for some function $\Phi$.
            \label{theo:gin}
        \end{theorem}
        Also, the universal approximation theorem \citep{univApprox} states that a multi-layer perceptron (MLP) with sufficiently large hidden dimensionality (i.e. from 100 to 500 nodes) and appropriate non-linearity (e.g. ReLU activation) can approximate any continuous function with an arbitrary accuracy. The Graph Isomorphism Network (GIN), developed based on the aforementioned theorems \citep{gin}, is one of the most expressive GNNs in the class of message-passing GNNs, where its expressiveness is upper bounded by Weisfeiler-Lehman Isomorphism Test. It has the following relation:
        \begin{equation}
            {\gls{hv}}_i^{(k+1)} = \text{MLP}^{(k)} \bigg( {\gls{hv}}_i^{(k)}, \text{Sum}^{(k)} \left(\{ {\gls{hv}}_j^{(k)} | \forall \gls{v}_j \in \gls{N}(\gls{v}_i)\} \right) \bigg)
            \label{equ:basic-mess-pass3}
        \end{equation}
        
        Therefore, sum pooling (Sum) and MLP will be used in a generalized message-passing framework. Note that we can also add rich node features to improve the expressiveness of the GNNs.

    \subsection{Graph Networks}
        A generalized message-passing framework is known as a graph network (GN) \citep{gn}. The important aspect in graph networks is that, during message passing, in addition to node embeddings ${\gls{hv}}_i$, we generate the edge embeddings ${\gls{he}}_i$, as well as a global embedding \gls{hu} corresponding to the entire graph. This allows the framework to easily integrate edge- and graph-level features. Also, recent work \citep{Barcel2020TheLE} has shown GN to have benefits in terms of expressiveness compared to a standard GNN. Generating edge and global embeddings during message-passing also makes it trivial to define loss functions based on graph or edge-level classification tasks \citep{gnnbook}.
        
        
        
        As already discussed, multi-layer perceptrons (MLPs) and element-wise sum pooling (Sum) are selected as the Update and Aggregate functions for maximal expressiveness. The message-passing neural network (MPNN, also called interaction network) architecture \citep{Gilmer2017NeuralMP}, a simplified version of the full GN architecture, will be used for the physics simulator explained in \S \ref{subsec:learning}.
        

    \subsection{Limitation and Proposed Approach}
        GNNs might be infeasible for large-scale applications on a GPU with mid-level limits on memory (i.e. VRAM of 10-20GB). Aside from mini-batch, one solution to this is to consider linear activation functions in GNNs. This has worked well for node classification benchmark \citep{wu2019simplifying}. Using this, one can pre-process graph features on a CPU (with large memory capability) offline.
        
        However, this will limit the expressive power of GNNs due to the lack of non-linearity in generating the embeddings. In this research, we propose applying a non-learning, CPU-runnable dimensionality reduction method, PCA, to physics graph data in order to:
        \begin{enumerate}
            \setlength\itemsep{0em}
            \item Significantly reduce the size of graph data,
            \item Reduce the required message-passing steps to only one, and
            \item Eliminate the need for computationally expensive edge construction algorithms.
        \end{enumerate}

\section{Subspace Graph Network Simulator} \label{sec:subgns}
    \begin{figure}[!t]
        \centering
        \includegraphics[width=.75\textwidth]{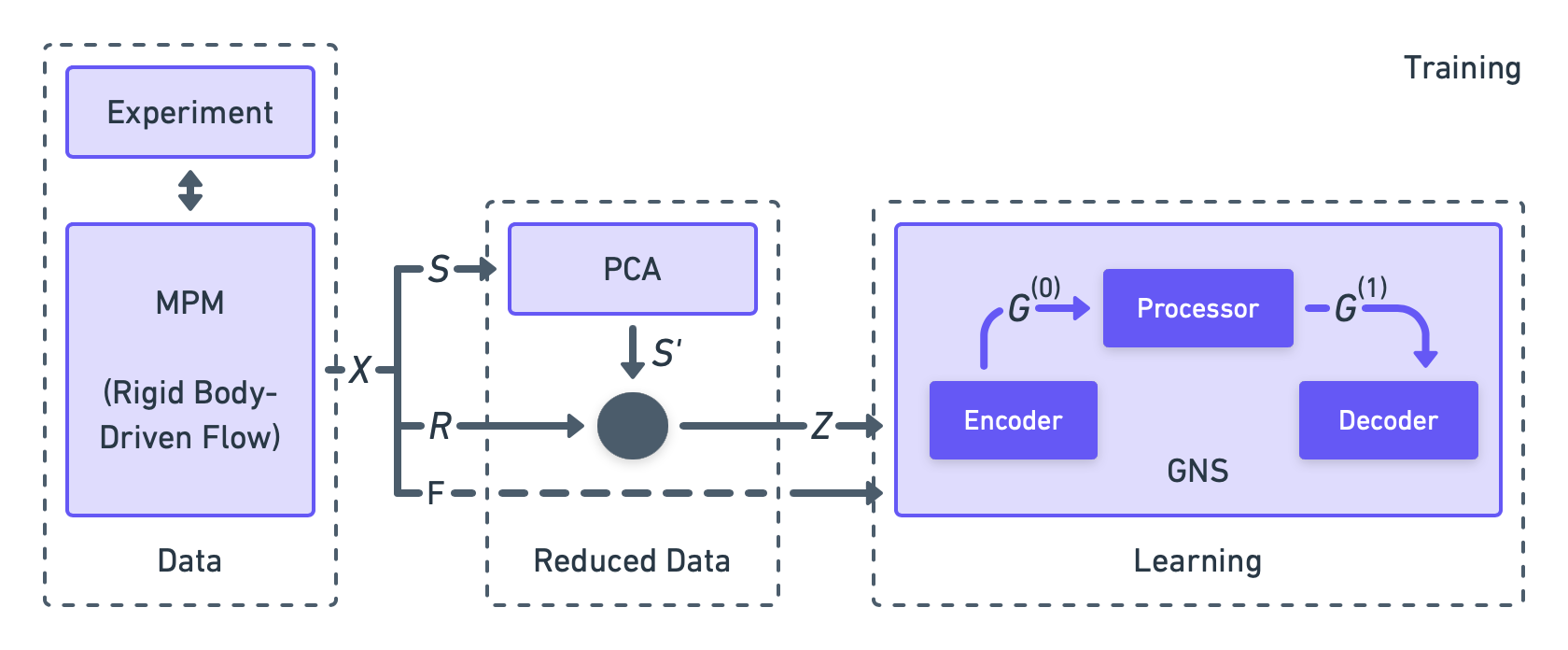}
        \caption{Training in subspace GNS. Data \gls{X} includes time series of flow states \gls{S}, rigid states \gls{R}, and interaction forces \gls{F}. Also, \gls{Sp} and \gls{Z} are reduced flow states and reduced system states. Here, states are particle positions.}
        \label{fig:dltraining}
    \end{figure}
    
   For engineering applications, the underlying physics govern the system response. A learning physics simulation can be discussed and assessed in terms of the following factors:
    \begin{enumerate}
        \setlength\itemsep{0em}
        \item Accuracy: An accurate method ideally learns from clean and accurate datasets (\S \ref{subsec:data}).
        \item Speed: A fast method can be achieved via reduced, memory-efficient data and/or by using reusable modules that can be run on modern accelerator hardware such as GPU (\S \ref{subsec:reduced}).
        \item Generalization: Strong generalization is achievable by imposing \textit{relational inductive bias} on the learning model. Models with strong inductive biases are more data efficient and can generalize much better to unseen scenarios (\S \ref{subsec:learning}).
        \item Differentiability: Differentiable models such as neural networks are appropriate for inverse problems (e.g. control systems).
    \end{enumerate}
    
    In this section, we develop a real-time learning simulation approach for modeling rigid body-driven elasto-viscoplastic (and sometimes stress-free) granular flows. Focus on this application does not limit its potential applicability to other materials, not considered here. The approach will benefit from material point method (MPM), principal component analysis (PCA), and graph network simulator (GNS), and is decomposed to a training phase and a rollout phase. Such an approach has not been developed for granular flows before.
    
    The training phase is illustrated in Figure \ref{fig:dltraining}. First, we generate training datasets via material point method (MPM) verified by experiments. Then, we apply PCA to the full data in a pre-processing step (on CPU). As a result, depending on the application, a desired quality (i.e. considering both accuracy and memory-efficiency) can determine the number of PCA modes. Finally, we train GNS (a GN model with Encoder-Processor-Decoder scheme on GPU) using data representing subspace elasto-viscoplastic granular flow. The modules shown in the figure including Data (\S \ref{subsec:data}), Reduced Data (\S \ref{subsec:reduced}), and Learning (\S \ref{subsec:learning}) will be elaborated in the following sections.
    
\begin{figure}[!t]
        \centering
        \includegraphics[width=.75\textwidth]{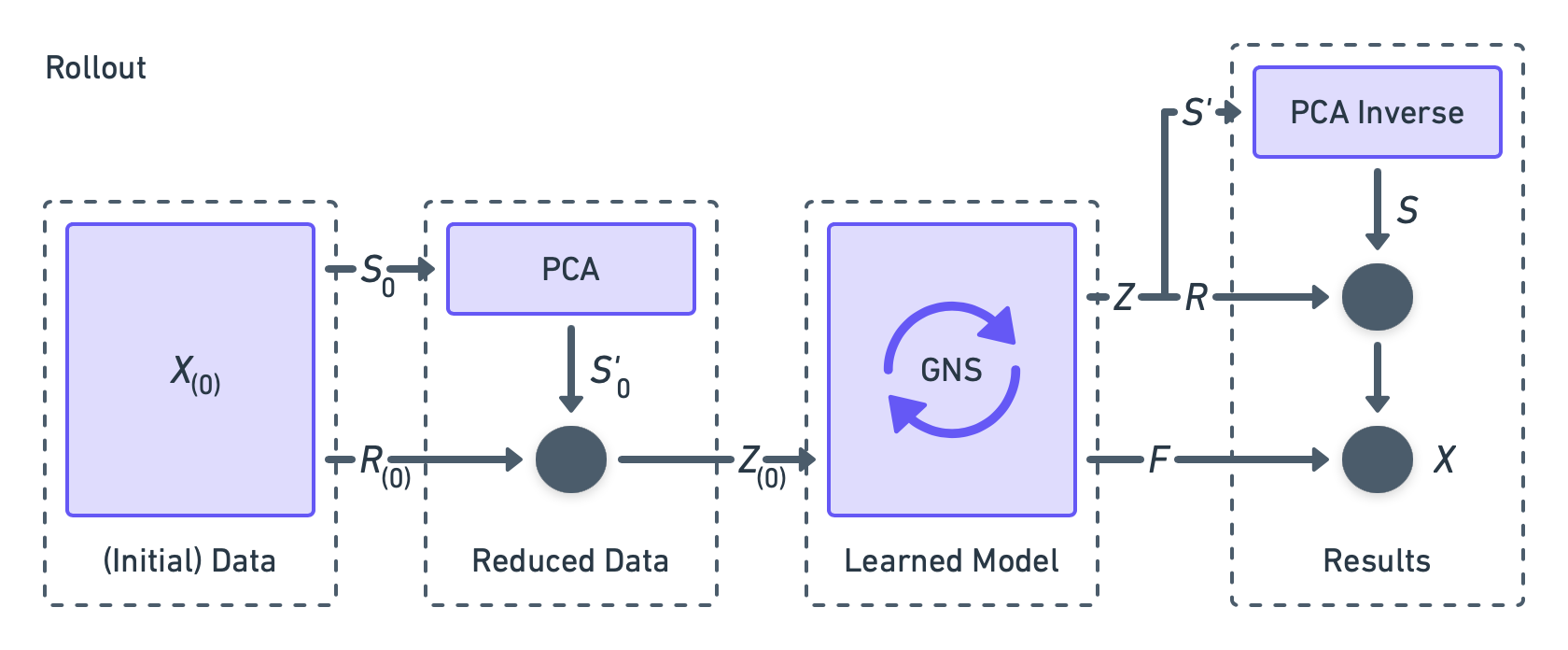}
        \caption{Rollout in subspace GNS. Initial Data $\gls{X}_{(0)}$ includes initial flow states $\gls{S}_0$ and (initial) rigid states $\gls{R}_{(0)}$. Also, $\gls{Sp}_0$ and $\gls{Z}_{(0)}$ are reduced initial flow states and reduced (initial) system states. In the end, results \gls{X} include time series of flow states \gls{S}, rigid states \gls{R}, and interaction forces \gls{F}. Also, \gls{Sp} and \gls{Z} are reduced flow states and reduced system states. Here, states are particle positions.}
        \label{fig:dlrollout}
    \end{figure}

    In the rollout phase, depicted in Figure \ref{fig:dlrollout}, we use the learned subspace model to compute the granular flow-rigid body interaction forces (on CPU). Also, the fullspace granular flows are computed in a post-processing step (on GPU) which is useful for visualization purposes. The rollout phase will be explained in \S \ref{subsec:rollout}.
    
    \subsection{Data} \label{subsec:data}
\begin{figure}[!t]
            \centering
            \includegraphics[width=1\textwidth]{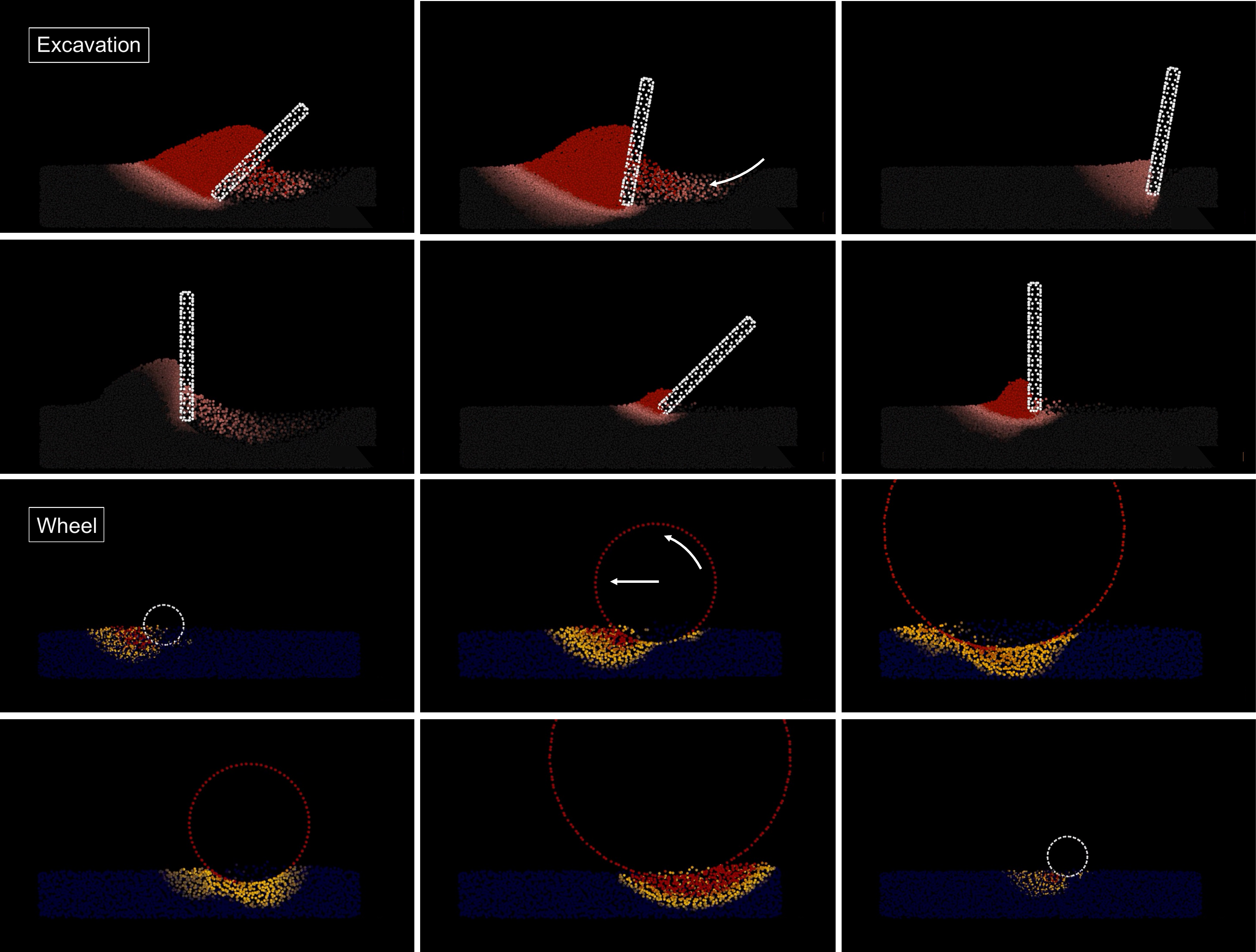}
            \caption{Side view (cross-section) of some Excavation and Wheel training examples with various initial and boundary conditions generated via MPM.}
            \label{fig:ds}
        \end{figure}

        Material point method (MPM) has been shown to provide good matching with experimental results \citep{Hae20isarc,Hae21aero} with runtime on the order of 10 to hundreds of times real-time. Therefore, hundreds of numerical simulations can be produced for the proposed machine learning approach. Here, we generate two datasets: Excavation and Wheel. Each dataset has $\sim 75000$ data frames gathered with 60 Hz of data acquisition rate (as temporal discretization is not constrained by stability conditions, in data-driven methods). The Excavation dataset contains 250 examples with a single soil type and a blade cutting at various depths, speeds, rake angles (i.e. relative to vertical), and motion types. The Wheel dataset is slightly more complex. This includes 243 examples with soils having different internal friction angles, and wheels with various diameters operating at various normal load and slip (percentage of the wheel's rotary motion not translating to forward linear motion). There are also multiple gravity conditions representing the Moon, Mars, and Earth. These variables are summarized in Table \ref{tab:ds}. Moreover, some examples are shown in Figure \ref{fig:ds}.
        
        \begin{table}[!t]
            \small
            \caption{Datasets specifications. Motion types include (1) a discrete ramp-horizontal-ramp motion, and (2) a continuous curved motion. Also, Grav., Fric., and Dia. stand for gravity, soil internal friction angle, and wheel diameter.}
            \begin{center}
            \begin{tabular}{lccccc}
            \toprule
            \textbf{Dataset} &\multicolumn{5}{c}{\textbf{Variables}}\\
            \toprule
            \\[-0.5em]
            \multirow{2}*{Excavation} &Angle [deg] &Depth [cm] &Speed [cm/s] &Motion\\
            \cline{2-5}
            &\{0,4,10,31,45\}&\{2,4,5,8,10\}&\{1,4,8,10,15\}&\{1,2\}\\
            \\[-0.5em]
            \hline \hline
            \\[-0.5em]
            \multirow{2}*{Wheel} &Grav. [$\text{m}/\text{s}^2$] &Fric. [deg] &Load [N] &Dia. [cm] &Slip [\%]\\
            \cline{2-6}
            &\{1.62,3.72,9.81\}&\{30,37,43\}&\{100,164,225\}&\{5,15,30\}&\{20,40,70\}\\
            \\[-0.5em]
            \bottomrule
            \end{tabular}
            \label{tab:ds}
            \end{center}
        \end{table}
    
        A dataset $\gls{X} = \{\gls{X}_i\}_{i=1:\gls{c}}$ includes \gls{c} examples of $\gls{X}_i = \{\gls{S}_i, \gls{R}_i, \gls{F}_i\}$ consisting of a time series of the positions of granular flow  $\gls{S}_i$ and the rigid body $\gls{R}_i$ interacting with it, and a time series of the total interaction forces $\gls{F}_i$ (applied to the center of mass of the rigid body), where 
        \begin{equation}
            \gls{S}_i =
            \begin{bmatrix}
            x_1^1 &y_1^1 &z_1^1 &\ldots &x_1^m &y_1^m &z_1^m \\
            x_2^1 &y_2^1 &z_2^1 &\ldots &x_2^m &y_2^m &z_2^m \\
            &\vdots & &\ddots & &\vdots \\
            x_n^1 &y_n^1 &z_n^1 &\ldots &x_n^m &y_n^m &z_n^m \\
            \end{bmatrix}^T_{3\gls{m} \times \gls{n}}
            \label{equ:data1}
        \end{equation}
        with \gls{m} and \gls{n} as the numbers of timesteps (frames) and flow particles, respectively. In fact, each column in $\gls{S}_i$ represents a time series of particle positions in 3D. The $\gls{R}_i$ has also a similar structure. Moreover, the interaction forces $\gls{F}_i$ are given by
        \begin{equation}
            \gls{F}_i =
                \begin{bmatrix}
                F_x^1 & F_x^2 & \ldots & F_x^m \\
                F_y^1 & F_y^2 & \ldots & F_y^m \\
                F_z^1 & F_z^2 & \ldots & F_z^m
                \end{bmatrix}^T_{\gls{m} \times 3}.
            \label{equ:data2}
        \end{equation}

        We divide the dataset into two subsets: the training split including 90\% and test split including 10\% of the examples. We work with the training split in the training phase, and with the test split in the rollout phase. Also, a validation split is not required as we will use the model hyperparameters already tuned.

        Note that as we separately construct the $\gls{S}_i$ and $\gls{R}_i$ data matrices, the particle types (i.e. flow, rigid/boundary, etc.) are recognizable. They are used as node features in the graph network explained in \S \ref{subsec:learning}.

    \subsection{Reduced Data} \label{subsec:reduced}
        Physical systems in MPM can be described by particles each with three coordinates (degrees of freedom) in 3D. The minimum number of particles are subject to stability conditions. Thus, such systems are often high-dimensional for large scale prolems. Note, to conserve the particle coordinates, we define the dimensions by the particles rather than the system's degrees of freedom. However, the effective dimensions of a physical system can be far smaller than the system dimensionality.
        
        Principal Component Analysis (PCA) is a non-parametric linear dimensionality reduction method \citep{pcaref1, pcaref2}. It provides a data-driven, hierarchical coordinate system to re-express high-dimensional correlated data. The resulting coordinate system geometry is determined by principal components. These are sometimes called bases, modes or latent variables of the data. These modes are uncorrelated (orthogonal) to each other, but have maximal correlation with the \textit{observations} \citep{pcabook}. Particularly, the goal of this method is to compute a subset of ranked principal components to summarize the high-dimensional data while retaining trends and patterns.
        PCA, as a model reduction method, can capture the effective dimensions of the physicals system. In fact, PCA enables us to calculate the complexity of the system. It should be noted that both the strength and weakness of PCA is that it is a non-parametric method. In cases where standard PCA fails due to the existence of non-linearly in data, kernel PCA \citep{kernelpca} can be used. Compared to an autoencoder neural network, kernel PCA still does not require learning/optimization and prior specification of size of the latent space.

        \textit{Reduced Data Computation:}
            We aim to apply PCA to the granular flow data, and pass the rigid body as is (\citep{pcamg}). Therefore, we compute the PCA transformation matrix $\gls{U}_{\gls{n} \times \gls{r}}$
            where \gls{r} is the number of modes we select, and the data matrix is $\gls{S}_\text{train}$ including the flow states of each example in the training split as follows
            \begin{equation}
                \gls{S}_\text{train} =
                \setcounter{MaxMatrixCols}{12}
                \begin{bmatrix}
                    \gls{S}_1 & \gls{S}_2 & \ldots & \gls{S}_{\gls{c}_\text{train}}
                \end{bmatrix}^T_{3 \gls{c}_\text{train} \gls{m} \times \gls{n}},
                \label{equ:reddata1}
            \end{equation}
            where $\gls{c}_\text{train}$ is the number of examples in the training split. Now we can compute the reduced flow states $\gls{Sp}= \{\gls{Sp}_i\}_{i=1:\gls{c}}$ by applying the PCA transformation $\gls{U}_{\gls{n} \times \gls{r}}$ separately to each example in $\gls{S}= \{\gls{S}_i\}_{i=1:\gls{c}}$ as follows
            \begin{equation}
                \gls{Sp}_i=(\gls{S}_i - \gls{Sbar}_\text{train}) \gls{U}.
                \label{equ:reddata2}
            \end{equation}
            Finally, we construct the system states $\gls{Z} = \{\gls{Z}_i\}_{i=1:\gls{c}}$, where
            \begin{equation}
                \gls{Z}_i =
                    \begin{bmatrix}
                        \gls{Sp}_i & \gls{R}_i
                    \end{bmatrix}.
                \label{equ:reddata3}
            \end{equation}
            We omit the subscript $i$ hereinafter for simplicity, and decompose the states for each example, $\gls{Z}_i$, into multiple $(\gls{d}+1)$ timestep windows of states $\gls{Z}^{t-(\gls{d}-1), \ldots , t, t+1}$ as the inputs to the Learning module (in \S \ref{subsec:learning}). Note, the $\gls{Z}^{t+1}$ are as the target (or label) states. The $\gls{F}^{t+1}$ is also passed to the Learning module as the target interaction force.
        
\begin{figure}[!t]
                \centering
                \includegraphics[width=0.75\textwidth]{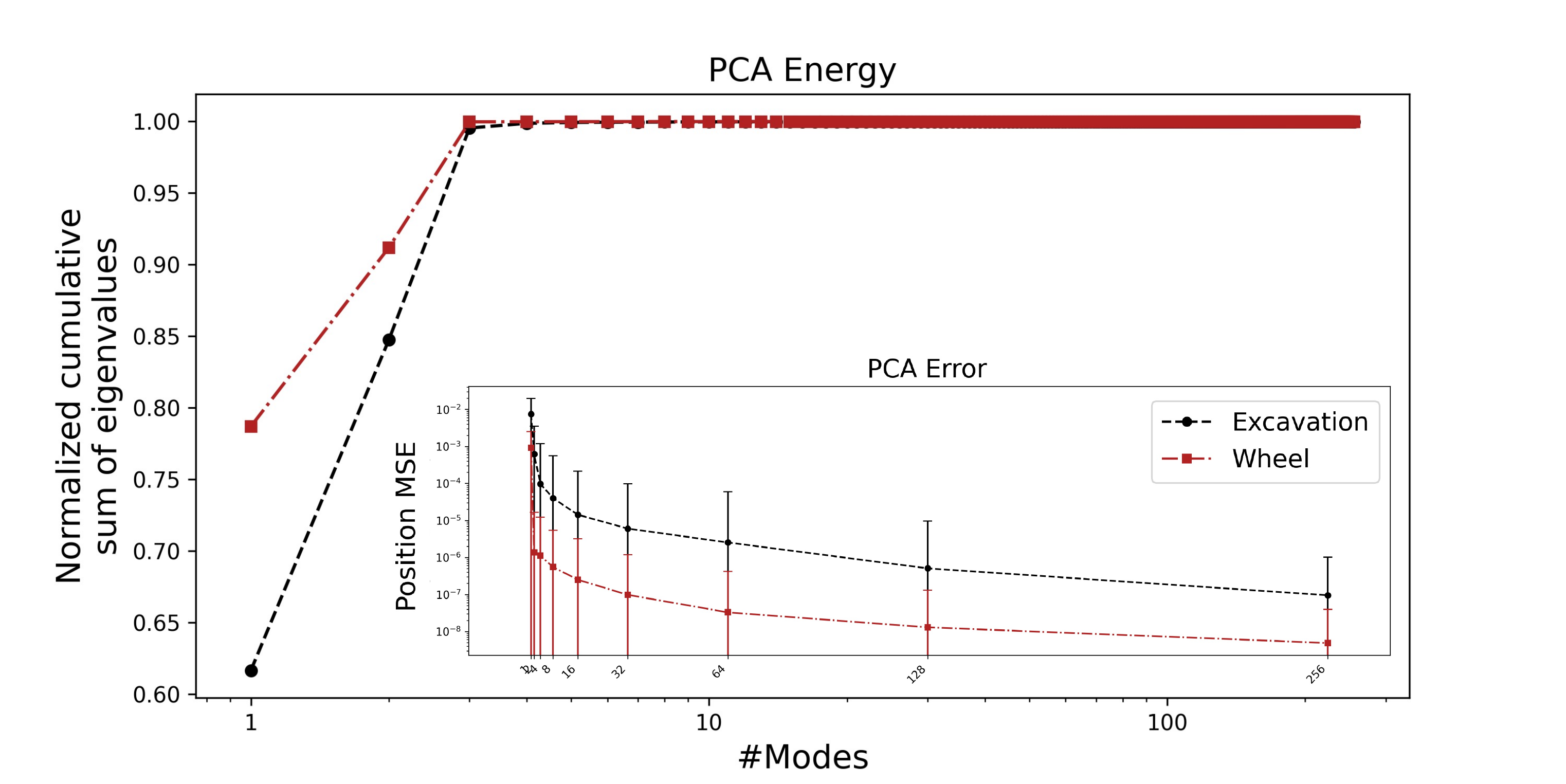}
                \caption{\textbf{Outset:} Normalized cumulative sum of the transformation matrix eigenvalues (modes) representing their energy. \textbf{Inset:} Position MSE between original full data and mapped-back full data. Errors bar represent one standard deviation.}
                \label{fig:pcaEnergyError}
            \end{figure}

        \textit{Mode Number Selection:}
            Each example in the Excavation and Wheel datasets is composed of $27000$ and $6000$ particles (or dimensions), respectively. To quantitatively determine the complexity of the physics data, we compute the normalized cumulative sum of the (sorted) eigenvalues corresponding to the columns (i.e. modes) of the transformation matrix \gls{U} via
            \begin{equation}
                \text{Energy} =
                    \frac
                    {\sum_{i=1}^{\gls{r}} \gls{lambda}_i}
                    {\sum_{i=1}^{\gls{n}} \gls{lambda}_i}.
                \label{equ:cumsumeig}
            \end{equation}
            This is used to select an appropriate number of modes, (\gls{r}), that can capture most of the Energy (i.e. variance) of the data \citep{pcabook}. Figure \ref{fig:pcaEnergyError}(outset) shows that the very first modes ($<10$ out of 27K and 6K dimensions) can be used to construct the Excavation and Wheel data in subspace (i.e. PCA space) while keeping more than $99\%$ of the data variance. Moreover, to measure the position mean squared error (MSE), we (1) map the data to the subspace using different numbers of modes ($r=1,2,4,\ldots,256$), (2) map back the reduced data to the fullspace, and (3) compare the mapped-back data with the original data. Based upon the results shown in Figure \ref{fig:pcaEnergyError}(inset), we choose 8 modes to use as inputs to the graph network (discussed in \S \ref{subsec:learning}) for learning the subspace dynamics of the granular flows. This will reduce the time and memory complexity of the graph network in both the training and rollout phases, while keeping position MSE below $10^{-4}$.
    
    \subsection{Learning} \label{subsec:learning}
        Graph neural networks (GNNs) should ``express'' as many \textit{inductive biases} as possible. Inductive biases allow learning algorithms to prioritize one solution over another, independent of the observed data \citep{indbias}. Also, \textit{relational reasoning} involves manipulating structured representations (graphs) of entities (nodes) and relations (edges), using rules (functions) for how they can be composed \citep{gn}. Graph networks (GNs) carry strong relational inductive biases which guide their approach towards learning about entities and relations \citep{gn}. They, in fact, impose constraints on relationships among graph nodes in a learning process. Furthermore, some non-relational inductive biases are used in deep learning, including non-linear activations, dropout \citep{dropout}, normalization \citep{batchnorm}, etc. all imposing constraints on the outcome of learning.
        
        The inductive biases can improve the \textit{combinatorial generalization} of GNs, an important aspect of a learning physics simulation approaches (factor \#3 listed in \S \ref{sec:subgns}). The principle of combinatorial generalization is constructing new predictions from known observation. This has been at the heart of the broader field of  artificial intelligence \citep{chomsky}. In addition, the GNs' node- and edge-level functions are shared across all edges and nodes, respectively. This means GNs automatically support a form of combinatorial generalization \citep{gn}.
        
        In terms of physics modeling, GNs have analogies with physical systems (in Newtonian mechanics) \citep{gnnphys, symreg}. If we assume that the graph nodes are like particles, then the particle internal forces, net internal forces, and accelerations, are analogous to the graph edges, Sum functions, and node Update function, respectively. These can be used iteratively to compute the positions in the next timestep. Here we aim to learn the subspace dynamics of the physical systems via graph network simulator (GNS) proposed by \citep{gns}. GNS has an Encoder-Processor-Decoder scheme which is a combination of the Encoder-Decoder and message-passing frameworks introduced in \S \ref{subsec:GNN}.
        
        GNS is a simulator $s_{\gls{theta}}: {\gls{Z}}^{t-(\gls{d}-1), \ldots , t-1, t} \to {\gls{Z}}^{t+1}$ with learning parameters \gls{theta}. It predicts the future states of a physical system $\hat {\gls{Z}}^{t+1}$ given a finite time history (\gls{d} timesteps) of the states ${\gls{Z}}^{t-(\gls{d}-1), \ldots , t-1, t}$. As mentioned in \S \ref{subsec:reduced}, \gls{Z} includes subspace particle positions and particle type information. GNS modules including Encoder, Processor, and Decoder will be described below along with minor but effective customization given our reduced data. Afterwards, the training specifications will be detailed.
        
        \subsubsection{Encoder}
            The $\text{Encoder}: \gls{Z}^{t-(\gls{d}-1), \ldots , t-1, t} \to \gls{G}^{(0)}$ embeds the particle states $\gls{Z}^{t-(\gls{d}-1), \ldots , t-1, t}$ as an initial latent graph $\gls{G}^{(0)}=(\gls{E}^{(0)},\gls{V}^{(0)})$ in the current timestep $t$. The $\gls{V}^{(0)}=\{{\gls{hv}}_i^{(0)}\}_{i=1:\gls{Nv}}$ is a set of node embeddings ${\gls{hv}}_i^{(0)}=\text{MLP}_{\text{enc},v} \big(\gls{v}_i^{(0)} \big)$ where $\gls{v}_i^{(0)}$ are node features. The $\gls{E}^{(0)}=\{({\gls{he}}_i^{(0)},\gls{ri},\gls{si})\}_{i=1:\gls{Ne}}$ is a set of edge embeddings ${\gls{he}}_i^{(0)}=\text{MLP}_{\text{enc},e} \big(\gls{e}_i^{(0)} \big)$ where $\gls{ri}$ is the index of receiver node, $\gls{si}$ is the index of sender node, and $\gls{e}_i^t$ are edge features. Note that the global feature \gls{u} (such as gravity and internal friction angle) has been omitted, as all required features can be incorporated into the node embeddings in our application.  
            
            Here, particle types and a finite time history of particle velocities ($\gls{d}-1$; velocities computed from particle positions) are considered as the node features $\gls{v}_i$. Also, relative positions between the sender and receiver nodes are added as the edge features $\gls{e}_i$. This choice of feature augmentation makes the position encoding spatial equivariant. In other words, this causes GNNs to perform better in position-aware tasks as GNNs have limited power in capturing the position of a given node with respect to all other nodes of the graph \citep{positionaware}. The general solution is known to construct the relative positions by anchoring a fixed point as reference. Here in GNS instead, the anchoring points are dynamic as they are the sender nodes.
            
            In the current approach, we construct a complete graph (i.e. all nodes are connected with all other nodes via two-way directed edges). This is because we are using physics particles as nodes in subspace and their relationships in this space might not be proximity-based (this is an open question for us). As a result, it eliminates the need to use a nearest neighbor algorithm for edge construction. 
                        
        \subsubsection{Processor.}
            The $\text{Processor}: \gls{G}^{(0)} \to \gls{G}^{(1)}$ is used as a message-passing step to compute the updated latent graph $\gls{G}^{(1)} = (\gls{E}^{(1)},\gls{V}^{(1)})$ given the initial latent graph $\gls{G}^{(0)}$ . This uses interaction network architecture \citep{Gilmer2017NeuralMP} -- a simplified version of the full GN architecture. The equations in the interaction network are shown in Algorithm \ref{algo:inet}.
            
            \begin{algorithm}[!t]
                \SetAlgoVlined
                \SetKwInOut{Input}{Input} 
                \SetKwInOut{Output}{Output}
                \ResetInOut{Output}
                \Input{$\gls{E}^{(0)}$, $\gls{V}^{(0)}$}
                \Output{$\gls{E}^{(1)}$, $\gls{V}^{(1)}$}
                \For{$i=1$ \KwTo \gls{Ne}}{
                    ${\gls{he}}_i^{(1)} \gets \text{MLP}_{\text{proc},e} \left({\gls{he}}_i^{(0)}, {\gls{hv}}_{\gls{ri}}^{(0)}, {\gls{hv}}_{\gls{si}}^{(0)} \right)$ \\
                }
                \For{$i=1$ \KwTo \gls{Nv}}{
                    $\bar{\gls{he}}_i^{(1)} \gets \text{Sum} \left(
                        \{ ({\gls{he}}_j^{(1)}, i, s_j) \}_{j=1:\gls{Ne}}
                    \right)$ \\
                    ${\gls{hv}}_i^{(1)} \gets \text{MLP}_{\text{proc},v} \left({\gls{hv}}_i^{(0)}, \bar{\gls{he}}_i^{(1)} \right)$ \\
                }
                \caption{Interaction network}
                \label{algo:inet}
            \end{algorithm}

            In addition to the spatial equivariance inductive bias (in the Encoder, already described above), the Processor introduces other physics-informed inductive biases including:
            \begin{enumerate}
                \setlength\itemsep{0em}
                \item Permutation equivariance: The Update functions (MLPs) are over a set of embeddings. The set representation does not impose ordering (whereas a vector does).
                \item Pairwise interactions: The edge Update function ($\text{MLP}_{\text{proc},e}$) captures the pairwise interactions between nodes (inter-particle forces in physics).
                \item Superposition principle: The sum pooling as the Aggregate function behaves similar to superposition (superposition principle in physics).
                \item Local interactions: The node Update function ($\text{MLP}_{\text{proc},v}$) captures the local interaction. It means that the effects of nearby nodes are larger than the farther nodes on a target node (nonlocality in physics).
                \item Universal rules: The node and edge Update functions (MLPs) are shared for each node and edge, respectively. Hence they can learn the underlying universal rules (the laws of physics).
            \end{enumerate}
            Also, another inductive bias (inertial motion) will be incorporated in the Decoder. These physics-informed inductive biases make GNS as general-purpose as possible for physics simulations. Note that, as a customization specific to modeling subspace physics, we use one message-passing step ($\gls{Nk}=1$) since our graph is complete and thus messages can reach any node from any other node in a single step.
            
        \subsubsection{Decoder}
            The $\text{Decoder}: \gls{G}^{(1)} \to \hat{\gls{o}}^{t+1}$ extracts the dynamics information $\hat{\gls{o}}^{t+1}$ from the updated latent graph $\gls{G}^{(1)}$. We define and compute the output $\hat{\gls{o}}^{t+1}$ via
            \begin{equation}
                \hat{\gls{o}}^{t+1} :=
                    \begin{bmatrix}
                        \hat{\gls{a}}_1^{t+1} & \ldots & \hat{\gls{a}}_{\gls{Nv}}^{t+1} \\
                        \hat{\gls{F}}_1^{t+1} & \ldots & \hat{\gls{F}}_{\gls{Nv}}^{t+1} 
                    \end{bmatrix}_{6 \times \gls{Nv}} =
                    \text{MLP}_\text{dec} \big(\gls{G}^{(1)} \big) 
                \label{equ:gnoutput}
            \end{equation}
            as a combination of particle (node) accelerations $\hat{\gls{a}}_i^{t+1}$ and particle interaction force contributions $\hat{\gls{F}}_i^{t+1}$ to the total interaction force $\hat{\gls{F}}^{t+1}=\sum_{i=1}^{\gls{Nv}} \hat{\gls{F}}_i^{t+1}$ applied to the center of mass of the rigid body. The $\text{MLP}_\text{dec}$ is applied to each particle separately. Also, the predicted particle states $\hat{\gls{Z}}^{t+1}$ can then be computed based on the predicted particle positions $\hat{\gls{z}}_i^{t+1}$ given by
            \begin{equation}
                \hat{\gls{z}}_i^{t+1} = {\gls{z}}_i^t + \Delta t \, \dot{\gls{z}}_i^t + \Delta t^2 \, \hat{\ddot{\gls{z}}}_i^{t+1}
                \label{equ:difequ}
            \end{equation}
            where $\dot{\gls{z}}_i^t = ({\gls{z}}_i^{t} - {\gls{z}}_i^{t-1})/ \Delta t$, $\hat{\ddot{\gls{z}}}_i^{t+1} = \hat{\gls{a}}_i^{t+1}$, and $\Delta t=1$.
            Basically, GNS focuses on the non-trivial, higher-order term (i.e. particle acceleration $\ddot{\gls{z}}$) of equation (\ref{equ:difequ}). In fact, this term is a function of the particle internal force \gls{f} (as $f=m\ddot{\gls{z}}$, with $m$ here being mass). And \gls{f} is a function of particle stress tensor \gls{T} including the effect of all elasto-visco-plastic (or sometimes stress-free) deformations (as $\gls{f} = \text{V} \text{div}\gls{T}$, with constant particle mass and volume, V; not to be confused with the set of nodes $V$). In GNS, it is also possible to include the effect of external forces by adding gravity, external friction coefficient, etc. to the node features. With these, GNS fundamentally becomes a force-based simulator like MPM by predicting the right hand-side of the momentum equation (as opposed to position-based methods like PBD).
            
            From another point of view, GNS predicts the deviation from the current position leading to a more stable prediction. Note that here, the accelerations are assumed to be constant between the timesteps. So the accelerations between timesteps $t$ and $t+1$ are shown by $\hat{\ddot{\gls{z}}}_i^{t+1}$ (or $\hat{\gls{a}}_i^{t+1}$). Also in equation (\ref{equ:difequ}), it is not necessary to include the $1/2$ coefficient since $\hat{\ddot{\gls{z}}}_i^{t+1}$ is the output of an MLP that can learn this coefficient.
            
        \subsubsection{Specifications}
            As explained above, five MLPs are used as the Update functions in GNS. These include $\text{MLP}_{\text{enc},e}$, $\text{MLP}_{\text{enc},v}$, $\text{MLP}_{\text{proc},e}$, $\text{MLP}_{\text{proc},v}$, and $\text{MLP}_\text{dec}$ whose parameters (i.e. weights) \gls{theta} should be optimized. Each MLP has two hidden layers (with ReLU activations) and an output layer each with size of 128, except for the $\text{MLP}_\text{dec}$ whose output layer size is 6 given our defined output vector (i.e. concatenated vector of particle acceleration and interaction force contribution).
            
            We define the loss function \gls{L} as the sum of the mean squared error (MSE) of the accelerations \gls{a} and squared error of the interaction force \gls{F} given by
            \begin{equation}
                \gls{L}(\gls{Z}^t, \gls{Z}^{t+1}; \gls{theta}) =
                    \bigg( \sum_{i=1}^{\gls{Nv}} ||\hat{\gls{a}}_i^{t+1} - \gls{a}_i^{t+1}||^2 \bigg)/\gls{Nv} +
                    ||\hat{\gls{F}}^{t+1} - \gls{F}^{t+1}||^2
                \label{equ:loss}
            \end{equation}
            where $\gls{F}^{t+1}$ is the target interaction force, and the target accelerations $\gls{a}_i^{t+1}$ are approximated using a temporal second-order central finite difference scheme on the positions given by
            \begin{equation}
                \gls{a}_i^{t+1} = \ddot{\gls{z}}_i^{t+1} = \frac{\gls{z}_i^{t-1}-2\gls{z}_i^{t}+\gls{z}_i^{t+1}}{\Delta t^2}
                \label{equ:fdscheme}
            \end{equation}
            where $\gls{z}_i^{t+1}$ are the target positions and $\Delta t=1$. Adam optimizer \citep{adam} is used to minimize the loss function over a mini-batch of size 2. The exponential learning rate is set to decay from $10^{-4}$ to $10^{-6}$ through a maximum of 20M training steps. We used the tuned model hyperparameters from \cite{gns}.

            \begin{figure}[!t]
                \centering
                \includegraphics[width=1\textwidth]{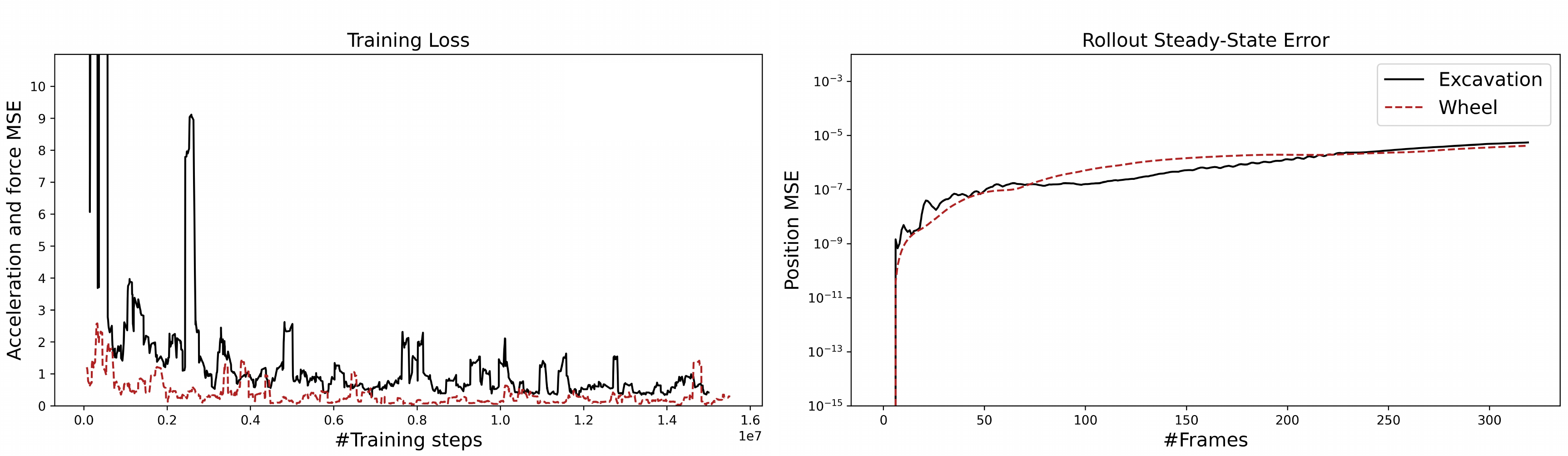}
                \caption{Training loss convergence in $\sim 15$M steps (left) and steady-state rollout error of positions for both Excavation and Wheel.}
                \label{fig:trrollerr}
            \end{figure}
            
            To accelerate the training process, the data is element-wise normalized to zero mean and unit variance with pre-computed means and variances of the training split. Also, a random-walk noise, $\text{Normal}(0,0.0003)$, is applied to the node positions and velocities \citep{gns}. This can resolve error accumulation issues; a roughly steady-state rollout Position MSE was achieved, as shown in Figure \ref{fig:trrollerr}(right), for both Excavation and Wheel. The training loss convergence achieved in $\sim 15$M steps is also shown in Figure \ref{fig:trrollerr}(left).
            
            The training was performed on an NVIDIA GeForce RTX 3080 10GB GDDR6X graphics card. The post-processing visualizations that will be described in \S \ref{subsec:rollout}, were done in Python. The open-source code of the GNS model has been developed primarily via TensorFlow \citep{tensorflow} and Graph Nets library \citep{gn} and published by Google DeepMind \citep{gns}. We have, however, customized it in a way to be feasible and appropriate for robotics applications by adding real-time 3D force prediction functionality. Our code is publicly available on GitHub \href{https://github.com/haeriamin/Subspace-Graph-Physics}{here\footnote{\href{https://github.com/haeriamin/Subspace-Graph-Physics}{github.com/haeriamin/Subspace-Graph-Physics}}}.

    \subsection{Rollout} \label{subsec:rollout}
        The training phase described in the previous section prepares the model for the rollout phase depicted in Figure \ref{fig:dlrollout}. The rollout consists of the Initial Data, pre-processing Reduced Data, Learned Model, and post-processing Results modules.
        
        Initial Data includes an initial \gls{d}-timestep history of flow states $\gls{S}_0$. Also, the rigid states should be included. The rigid states can be either initialized ($\gls{R}_0$) or scripted/interactive (\gls{R}). The former means that the model will predict the trajectory of the rigid body interacting with flow as well; this choice should be consistent with the model already trained. Next, in Reduced Data, the initial flow states is reduced to a subspace representation ($\gls{Sp}_0$) using PCA (see equation (\ref{equ:reddata2})). The flow and rigid states $\gls{Z}_{(0)} = \{\gls{Sp}_0, \gls{R}_{(0)} \}$ are then input to the learned GNS model. This model predicts reduced flow states \gls{Sp} and full interaction forces \gls{F} applied to the center of mass of the rigid body (and full rigid states \gls{R} if applicable). The reduced flow states can then be mapped back (\gls{S}) using PCA Inverse as follows
        \begin{equation}
            \gls{S} = (\gls{Sp} \, \gls{U}^T)  + \gls{Sbar}_\text{train}.
            \label{equ:roll1}
        \end{equation}
        For interactive applications, the pre-processing PCA and post-processing PCA Inverse steps can be performed using GPU, and the subspace GNS using CPU \citep{pcamg}. Note that here states include particle positions. 
        
        In the following, we will examine the efficiency and accuracy of our approach using an example in the test split of the Excavation and Wheel datasets. Then, we present some further results to compare the ground truths and predictions in subspace and over multiple examples in the test split.
        
            \begin{table}[!t]
                \small
                \caption{MPM (Physics), subspace GNS (Ours) and fullspace GNS (Ref) runtime comparison. GPU$^1$, GPU$^2$, CPU$^1$, and CPU$^2$ are NVIDIA GeForce RTX 3080 10GB, NVIDIA Tesla P6 16GB, Intel Core i7-10700 2.9GHz 8-Core, and Intel Core i7-6700 3.4GHz Quad-Core, respectively. One CPU$^1$ core and four CPU$^2$ cores are used.}
                \begin{center}
                \begin{tabular}{l||l||c|c||c|c||c}
                \toprule
                \multirow{3}*{\textbf{Dataset}} &\multirow{3}*{\textbf{Method}} &\multicolumn{2}{c||}{\textbf{Training}} &\multicolumn{2}{c||}{\textbf{Rollout}} &\textbf{Simulation} \\
                & &\multicolumn{2}{c||}{[sec/step]} &\multicolumn{2}{c||}{[sec/sec]} &[sec/sec] \\
                & &\textbf{GPU}$^1$ &\textbf{GPU}$^2$ &\textbf{CPU}$^1$ &\textbf{GPU}$^1$ &\textbf{CPU}$^2$ \\
                \hline \hline
                \multirow{3}*{Excavation} &Physics &-- &-- &-- &-- &270\\
                \cline{2-7}
                &Ref &OOM &1.000 &-- &OOM &--\\
                \cline{2-7}
                &Ours &0.016 &0.064 &0.350 &0.440 &-- \\
                \hline \hline
                \multirow{3}*{Wheel} &Physics &-- &-- &-- &-- &240\\
                \cline{2-7}
                &Ref &OOM &0.500 &-- &OOM &--\\
                \cline{2-7}
                &Ours &0.017 &0.068 &0.380 &0.460 &--\\
                \bottomrule
                \end{tabular}
                \label{tab:gns}
                \end{center}
            \end{table}
        \subsubsection{Efficiency}
            Table \ref{tab:gns} provides the runtime of our approach, subspace GNS, in training and rollout phases using CPU (with one core) and GPU (including TFRecord file deserialization). It also includes the fullspace GNS runtime as reference. Due to out-of-memory (OOM) issue, the fullspace GNS with our large-scale configurations was not trainable on the aforementioned GPU (RTX 3080 10GB). However, using an NVIDIA Tesla P6 16GB graphics card, we were able to measure the fullspace GNS runtime on the Excavation and Wheel datasets.
            The results show that our approach is real-time (also $\sim$700x faster than MPM) whereas the fullspace GNS is not even feasible for large-scale 3D physics configurations.

        \subsubsection{Accuracy}
            We separately assess the accuracy of PCA and GNS. To this end, we compare the distribution of PCA and GNS position MSEs in Excavation and Wheel configurations. Particularly, we compute the MSE of each particle's positions through time (using 8 PCA modes). To obtain the PCA error, we (1) apply PCA to the original MPM data, (2) apply PCA Inverse to the resulting reduced MPM data, and (3) compute the MSE between the resulting mapped-back data (i.e. PCA Inverse(PCA(MPM))) and original MPM data. To obtain the GNS error, we (1) perform GNS using the reduced MPM data (one example in test split), (2) apply PCA Inverse to the resulting predicted reduced GNS data, and the reduced data, and (3) compute the MSE between resulting mapped-back predicted data (i.e. PCA Inverse(predicted GNS); recall GNS works in the reduced subspace) and mapped-back data (i.e. PCA Inverse(PCA(MPM))). Thus they are separate errors.
            
            \begin{figure}[!t]
                \centering
                \includegraphics[width=1\textwidth]{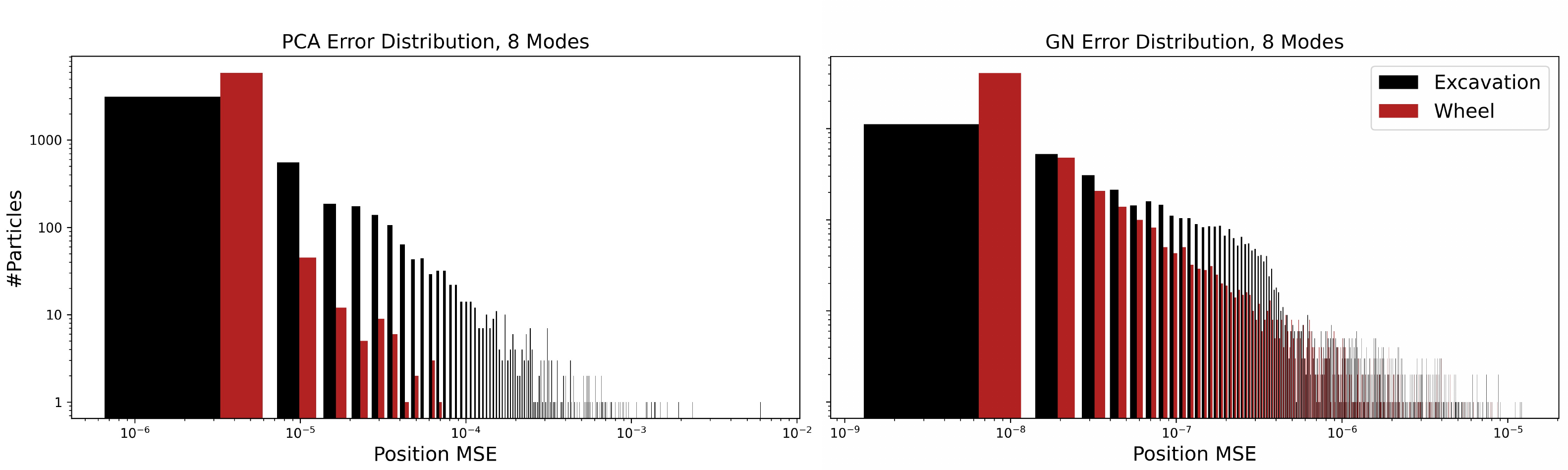}
                \caption{Distribution of PCA and GNS position MSEs in Excavation and Wheel configurations. Note that the varying width of bars is only due to logarithmic scaling of the x-axis.}
                \label{fig:pca_gn_error_dist}
            \end{figure}
            
            The histograms shown in Figure \ref{fig:pca_gn_error_dist} indicate that the position MSE ranges from $10^{-6}$ to $10^{-2}$ in PCA, and from $10^{-9}$ to $10^{-5}$ in GNS. But most particles have the lowest error in these ranges. This is somehow expected as most of the particles are stationary in both the Excavation and Wheel configurations. However, the notable point here is that the order of PCA error is higher than the GNS error. This suggests that a more accurate dimensionality reduction method with the same mode number can improve the accuracy of the final particle positions. The rest of the current section will focus on the accuracy of GNS in predicting the reduced data (i.e. in subspace learning).
            
            \begin{table}[!t]
                \small
                \caption{Test example specifications. Motion type 2 indicates blade has continuous curved motion. Also, Grav., Fric., and Dia. stand for gravity, soil internal friction angle, and wheel diameter.}
                \begin{center}
                \begin{tabular}{lccccc}
                \toprule
                \textbf{Dataset} &\multicolumn{5}{c}{\textbf{Variables}}\\
                \toprule
                \\[-1em]
                \multirow{2}*{Excavation} &Angle [deg] &Depth [cm] &Speed [cm/s] &Motion\\
                \cline{2-5}
                &45 &10 &10 &2\\
                \\[-1em]
                \hline \hline
                \\[-1em]
                \multirow{2}*{Wheel} &Grav. [$\text{m}/\text{s}^2$] &Fric. [deg] &Load [N] &Dia. [cm] &Slip [\%]\\
                \cline{2-6}
                &3.72 &30 &100 &15 &40 \\
                \\[-1em]
                \bottomrule
                \end{tabular}
                \label{tab:testexample}
                \end{center}
            \end{table}
            
            \begin{figure}[!t]
                \centering
                \includegraphics[width=1\textwidth]{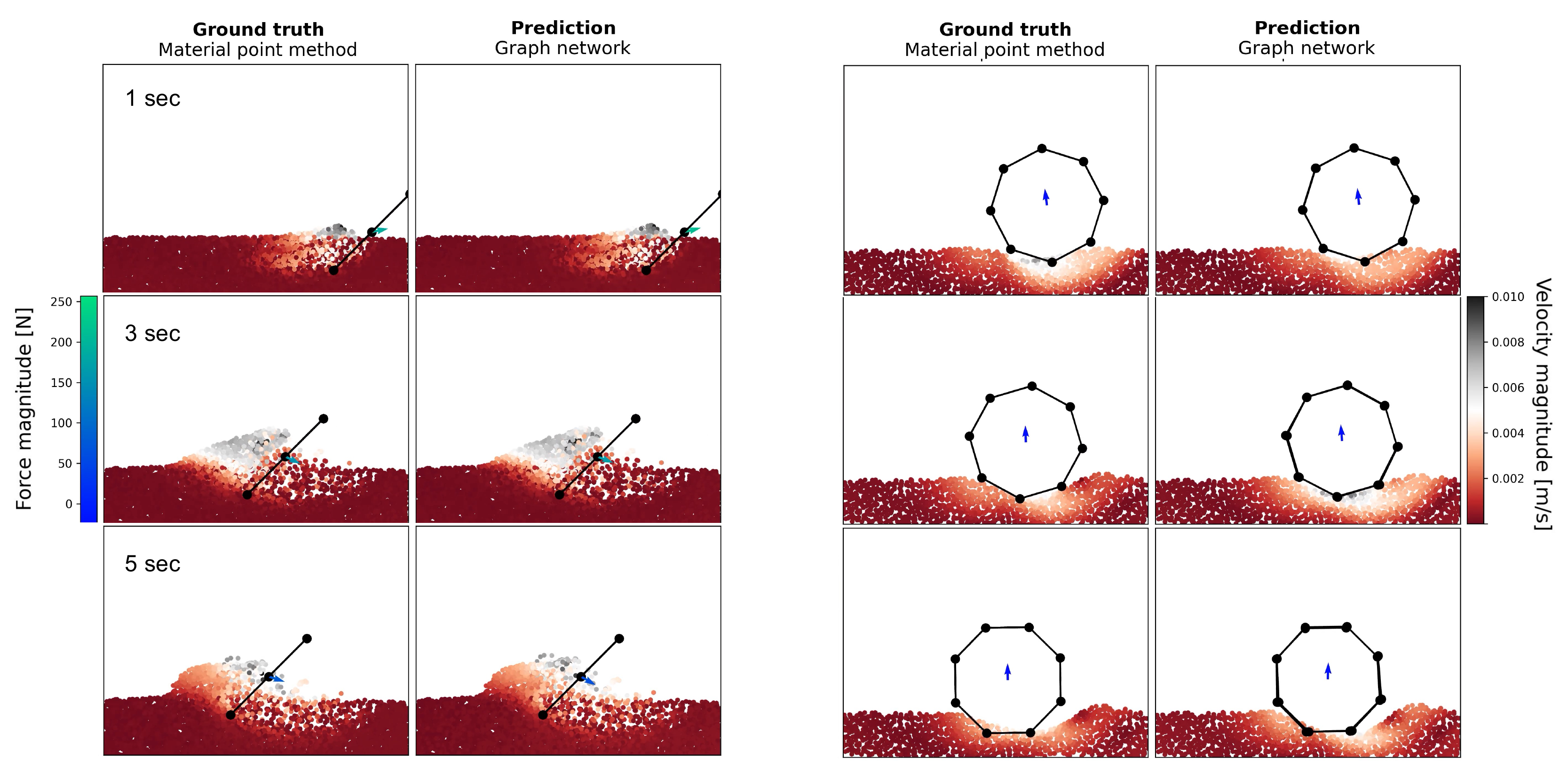}
                \caption{Side views of the Excavation and Wheel (cross-section) ground truth and predicted simulations in time (at 1, 3, and 5 second marks). Red to black colors indicate the particle velocity magnitudes. Arrows with blue to green colors represent directions and magnitudes of interaction forces.}
                \label{fig:gn_exwh2d}
            \end{figure}
            
            The specifications of each example chosen from the Excavation and Wheel test splits are shown in Table \ref{tab:testexample}. Note that in the Wheel configuration, the model learns the motion of the rigid body (i.e. wheel) as well. Also, the properties of flow (i.e. soil) and rigid body in the Wheel dataset are varying. These make the Wheel configuration more challenging for the GNS to learn.
            
            Now we assess the GNS accuracy in predicting particle positions and velocities in subspace. Figure \ref{fig:gn_exwh2d} shows side views of the Excavation and Wheel ground truth and predicted simulations in time (using the test examples introduced above). Red to black colors indicate the particle velocity magnitudes. There is a good agreement between the ground truth and predicted elasto-viscoplastic flows.
            
            \begin{figure}[t]
                \centering
                \includegraphics[width=1\textwidth]{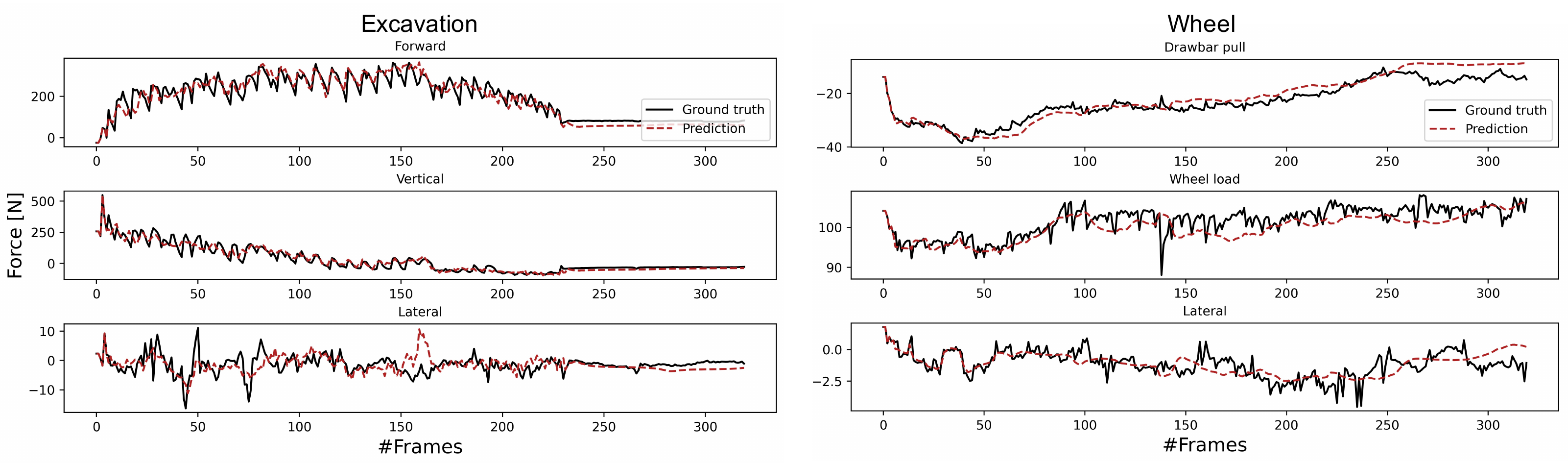}
                \caption{Ground truth and predicted Excavation and Wheel interaction forces. They are plotted in forward (drawbar pull), vertical (wheel load), and lateral directions.}
                \label{fig:gn_exwh_force}
            \end{figure}
            
            \noindent Note that the results have been mapped back by PCA Inverse. Also, arrows with blue to green colors represent the directions and magnitudes of interaction forces applied to the center of mass of the blade or wheel.
            
            We further assess the GNS accuracy in predicting interaction forces in subspace through time. It should be noted that the PCA accuracy does not directly affect the GNS force prediction accuracy as the force prediction is done in subspace via GNS. In other words, as long as the GNS learns the subspace dynamics accurately the force prediction will be accurate as well. Figure \ref{fig:gn_exwh_force} shows the ground truth and prediction interaction forces in the aforementioned test examples of the Excavation and Wheel datasets. The GNS model has predicted the Excavation forces more accurately than the Wheel forces (see Table \ref{tab:testexamples}). As mentioned the Wheel dataset is more challenging to learn. Therefore, we suggest training multiple Wheel GNS models separately by datasets each with the same soil properties and/or gravity condition (like the Excavation dataset).
            
            Finally, we summarize each example's position and force prediction errors in Table \ref{tab:testexamples}. It includes position MSE and force MPE (mean percentage error) of the 9 examples in the test split. Also, MPM force MPEs (relative to experiment) are provided for reference. Forces are in forward and vertical directions in the Excavation and Wheel configurations, respectively. As already discussed, force prediction is more accurate in the Excavation dataset due to no variation in the soil and blade properties in this dataset (i.e. only the blade motion is varying). Also, because of higher motion in the Excavation dataset, the mean position MSE is higher in Excavation than in Wheel. Additional results are presented in Appendix \ref{app:additional}.
            
            \begin{table}
                \small
                \caption{Position MSE and force MPE of test examples in subspace GNS rollout. MSEs (mean squared error) and MPEs (mean percentage errors) are specified relative to MPM. Also, MPM force MPEs are relative to experiment. Forces are in forward and vertical directions in Excavation and Wheel respectively.}
                \begin{center}
                \begin{tabular}{c||c|c||S[table-format=3.1]|S[table-format=3.1]}
                \toprule
                \textbf{Test} &\multicolumn{2}{c||}{\textbf{Position MSE} $\times 10^{-4}$} &\multicolumn{2}{c}{\textbf{Force MPE} [\%]} \\
                \textbf{Example} &{Excavation} &{Wheel} &{Excavation} &{Wheel} \\
                \hline \hline
                1 &0.01 &0.01 &4.5 &10.9 \\
                2 &0.01 &0.02 &-5.3 &-16.2 \\
                3 &0.00 &0.00 &0.3 &-18.5 \\
                4 &0.02 &0.03 &-10.5 &-22.8 \\
                5 &2.41 &0.05 &11.6 &-21.7 \\
                6 &0.02 &0.09 &-2.4 &-5.1 \\
                7 &2.86 &0.03 &-17.2 &0.2\\
                8 &0.01 &0.02 &-1.9 &26.6 \\
                9 &0.03 &0.03 &-5.1 &-31.0 \\
                \hline
                \textbf{Mean} &0.60 &0.03 &\textemdash &\textemdash \\
                \hline \hline
                \textbf{MPM} &\textemdash &\textemdash &-0.5 &5.2 \\
                \bottomrule
                \end{tabular}
                \label{tab:testexamples}
                \end{center}
            \end{table}

\section{Conclusion}
    This research considered the advancement of a subspace machine learning simulation approach. To generate training datasets, we utilized our high-fidelity (non-real-time) continuum method comprising MPM and NGF verified by experiments. Principal component analysis (PCA) was used to reduce the dimensionality of data. This showed that the first few principal components ($<10$ out of thousands of dimensions) can be used to construct the physics data in subspace while keeping more than $99\%$ of the data variance. A graph network simulator (GNS) was trained to learn the underlying subspace dynamics. In subspace GNS, we constructed a complete graph due to the fact that the relationship between physics particles as nodes in subspace might not be proximity-based. The complete graph allowed us to use only one message-passing step, and to omit the use of a time-consuming nearest neighbor algorithm for edge construction. The learned subspace GNS was then able to predict subspace particle accelerations and fullspace interaction forces with good accuracy and roughly steady-state rollout error (position MSE below $10^{-4}$). Force prediction was more accurate in the Excavation dataset due to no variation in the flow and rigid body properties in this dataset. More importantly, PCA significantly enhanced the time and memory efficiency of GNS in both training and rollout. That is, using 8 principal components enabled GNS to be trained using a single desktop GPU with moderate 10GB of VRAM. This also made the GNS real-time on large-scale 3D physics configurations (i.e. 700x faster than our continuum method which were used to produce the datasets).
    
    Although significant progress has been made in accurate and real-time simulation of rigid-body driven granular flows, several improvements can still be made for more accurate and efficient approaches including:
    
    \begin{enumerate}
        \setlength\itemsep{0em}
        \item An autoencoder can be used to reduce the dimensionality of data while capturing non-linearity in data. However, it requires learning procedure and an \emph{a priori} defined size of latent space. These are important limiting factors for high-dimensional data using limited-memory GPU hardware. As an alternative, kernel PCA can be employed to capture non-linearity in data without the above-mentioned restrictions \citep{kernelpca}.
    
        \item Since particle relations might not be proximity-based in subspace, a complete graph is constructed in subspace GNS. However, it may be more memory-efficient to find the potential relations between the selected principal components and construct the effective edges e.g. using the neural relational inference (NRI) model proposed by \cite{kipf2018neural}. It can be even more computationally efficient if one can find an analytical relation between the (sorted) principal components (as the time complexity of GNS edge update is proportional to edge number, $O(\gls{Ne})$).
    
        \item The existence of subspace dynamics has recently been discussed in \citep{SINDy}. Also, a combination of fullspace GNS and symbolic regression has recently been proposed to extract dynamics equations using pure data \citep{symreg}. So such approaches can be used to capture the potential subspace analytical governing equation for further analysis, and for potentially even better generalization than subspace GNS \citep{symreg}.
    
        \item More research should be done to explore the generalization of subspace GNS. For instance, we might be able to (1) train the subspace GNS using a specific number of principal components, and then (2) perform rollouts with different number of principal components depending on the accuracy required. 
    
        \item To potentially further reduce the GNS steady-state rollout error, a loss function over a finite next time horizon can be used as suggested by \citep{pcamg}.
    
        \item Combining GNS with numerical solvers and/or physics-informed loss functions will also be a promising direction for even better generalization in specific applications. Such approaches have already been developed with classical deep learning architectures \citep{physinformed, physinformed1, physinformed2, deeponet}.
    \end{enumerate}

\section{Acknowledgements}
    The authors gratefully acknowledge funding for this work from the Natural Sciences and Engineering Research Council of Canada (RGPIN-2015-06046), as well as collaboration with CM Labs Simulations Inc.

\bibliographystyle{cas-model2-names}
\bibliography{refs}
\clearpage

\appendix
\numberwithin{equation}{section}
\numberwithin{figure}{section}
\numberwithin{table}{section}
\section{Additional Results} \label{app:additional}
\begin{figure}
        \centering
        \includegraphics[width=1\textwidth]{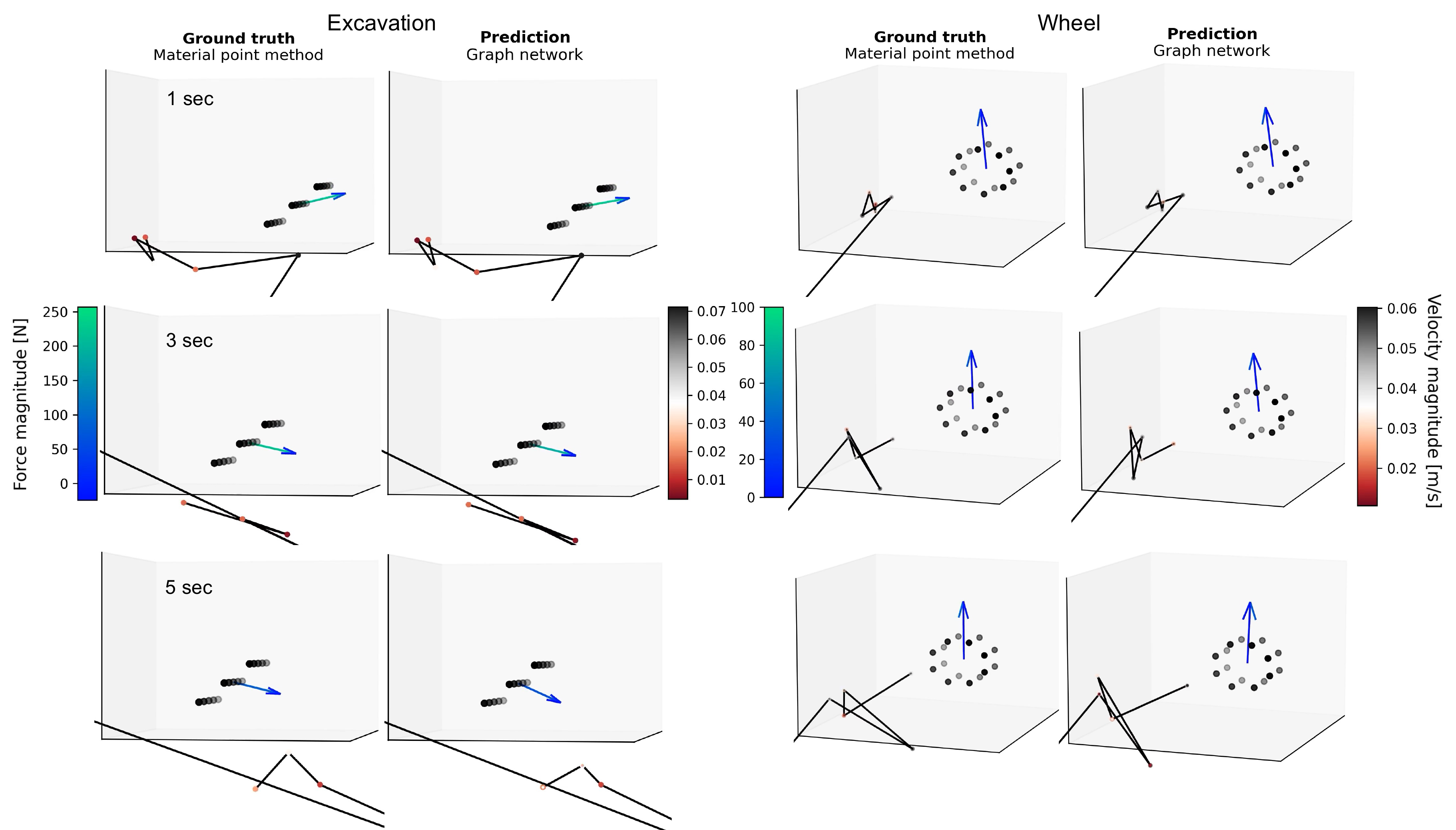}
        \caption{Underlying subspace dynamics predicted by GNS in time for both the Excavation and Wheel configurations (vs ground truth). Subspace particle velocity magnitudes and fullspace interaction forces are visualized.}
        \label{fig:gn_exwh3d_sub}
    \end{figure}

    Here, we first visualize GNS subspace prediction in 3D. Figure \ref{fig:gn_exwh3d_sub} reveals the underlying subspace dynamics predicted by GNS in time (at 1, 3, and 5 second marks) for both the Excavation and Wheel configurations. The subspace velocities of the 8 particles as well as the fullspace interaction forces are shown. This demonstrates the existence of dynamics in subspace that GNS has been able to learn. Such subspace dynamics have also been discussed in \citep{SINDy}. Methods such as symbolic regression can be used to extract the potential subspace parametric governing equation for further analysis \citep{symreg}.
    
\begin{figure}
        \centering
        \includegraphics[width=0.8\textwidth]{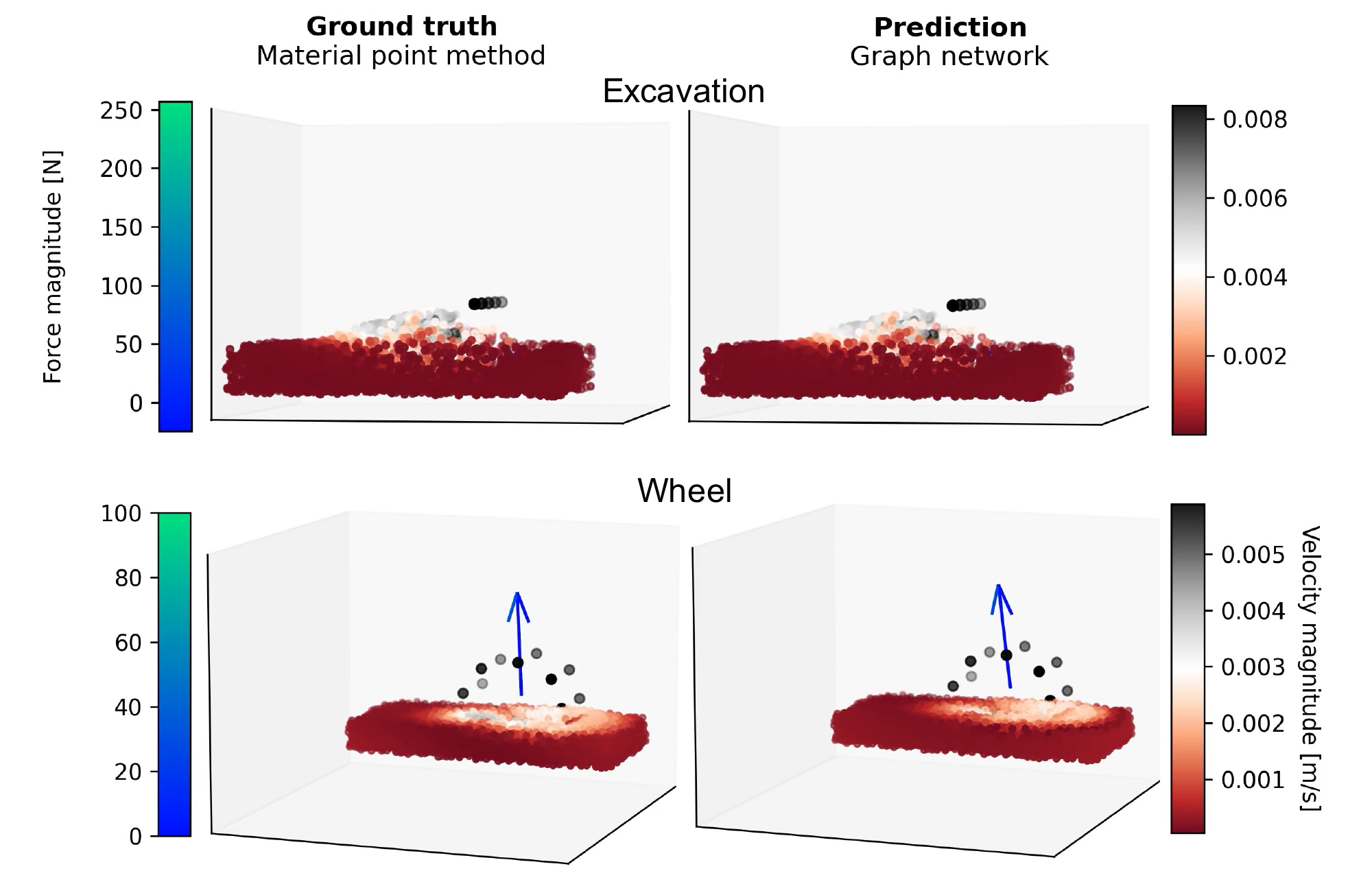}
        \caption{Underlying fullspace dynamics predicted by GNS at 3 second mark for both the Excavation and Wheel configurations (vs ground truth). Fullspace particle velocity magnitudes and interaction forces are visualized.}
        \label{fig:gn_exwh3d}
    \end{figure}

    In addition, the 3D mapped-back results (by PCA Inverse) are shown in Figure \ref{fig:gn_exwh3d}. As already explained and shown in these figures, the forces themselves are not reduced by PCA. The forces are not inputs to the GNS model; they are predicted based on the learned subspace dynamics.
    
\begin{figure}
        \centering
        \includegraphics[width=1\textwidth]{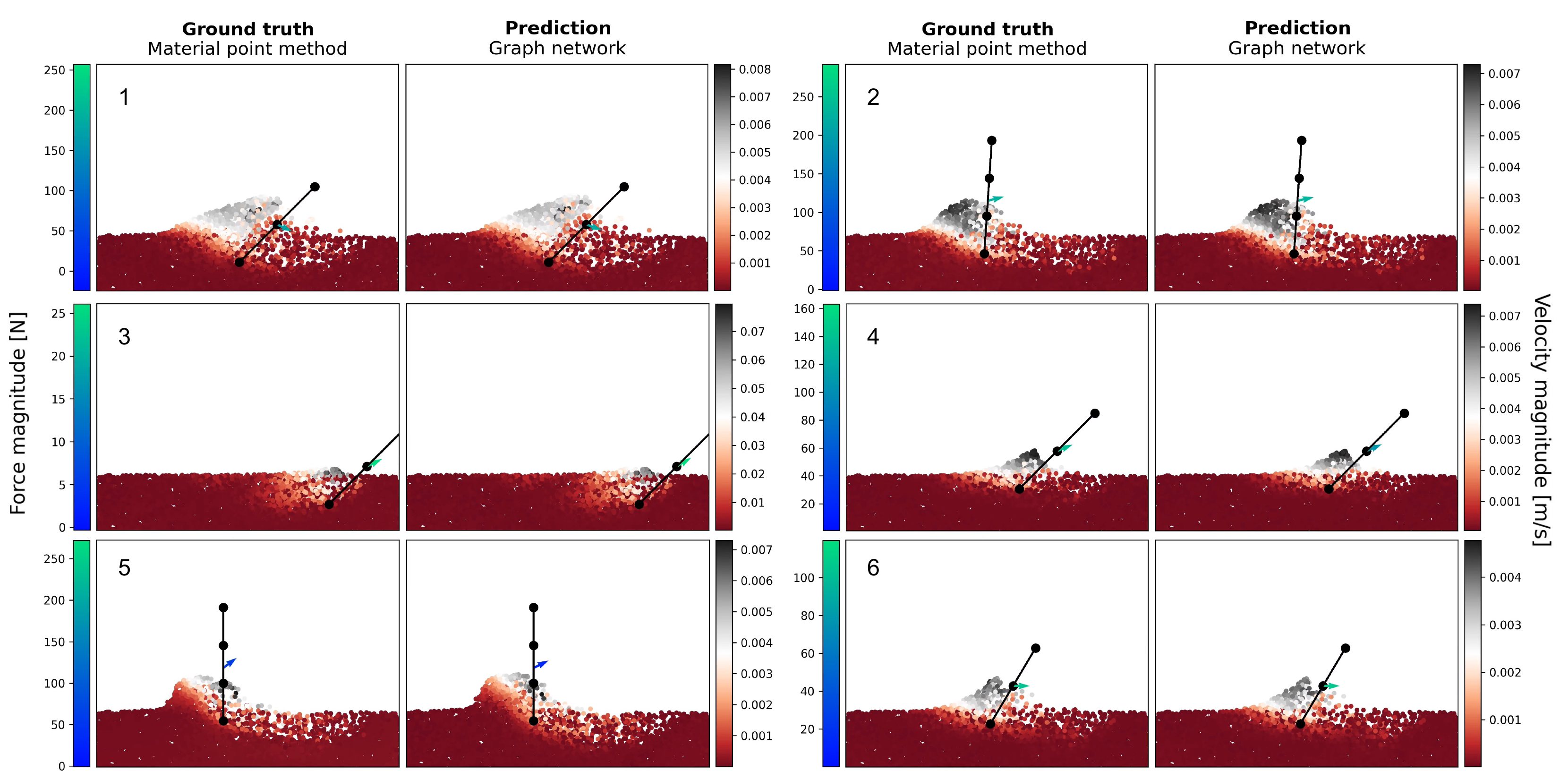}
        \caption{Side views of 6 Excavation test examples at 3 second mark. Red to black colors indicate the particle velocity magnitudes. Arrows with blue to green colors represent directions and magnitudes of interaction forces.}
        \label{fig:gn_excav_test}
    \end{figure}

\begin{figure}
        \centering
        \includegraphics[width=1\textwidth]{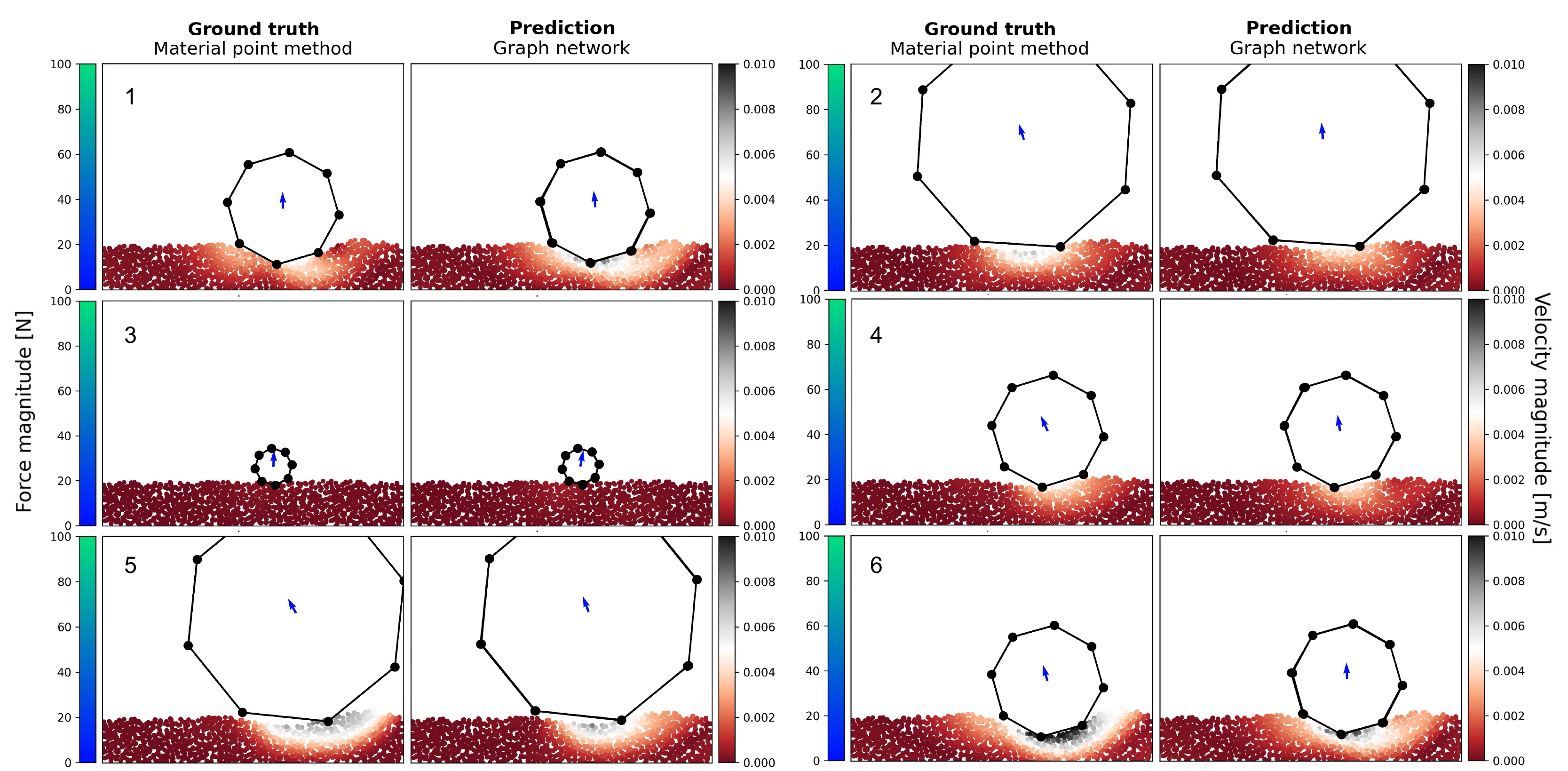}
        \caption{Side views of 6 Wheel test examples at 3 second mark. Red to black colors indicate the particle velocity magnitudes. Arrows with blue to green colors represent directions and magnitudes of interaction forces.}
        \label{fig:gn_wheel_test}
    \end{figure}

    We also provide the results of multiple examples in test splits. Figures \ref{fig:gn_excav_test} and \ref{fig:gn_wheel_test} show the mapped-back particle positions and velocities, and interaction forces at 3 second mark of 6 test examples in Excavation and Wheel datasets, respectively. As observed in the Excavation results, we can even change the discretization of the rigid body and still achieve accurate results (there are 20 rigid particles in number 2 and 4 whereas 15 in others). Although this should be examined further, in the current research this could be an indication that GNS models the particle interactions similarly to how a physics simulation approach such as MPM does. More research on this has been detailed in \citep{gns}.
    
\begin{figure}
        \centering
        \includegraphics[width=0.95\textwidth]{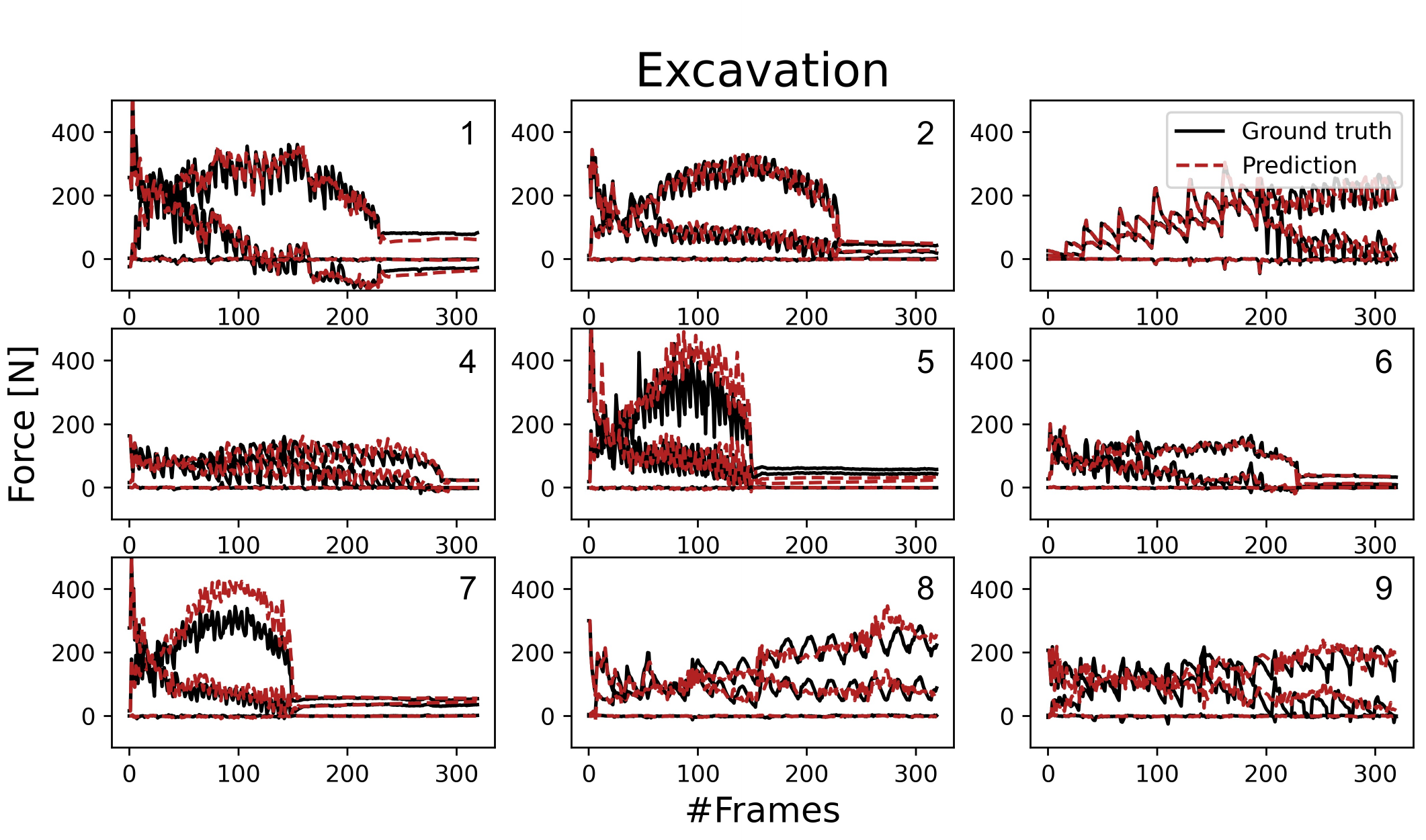}
        \caption{Interactions forces in 9 Excavation test examples. Each plot contains forces in forward, vertical and lateral directions.}
        \label{fig:forces_excav_test}
    \end{figure}

\begin{figure}
        \centering
        \includegraphics[width=0.95\textwidth]{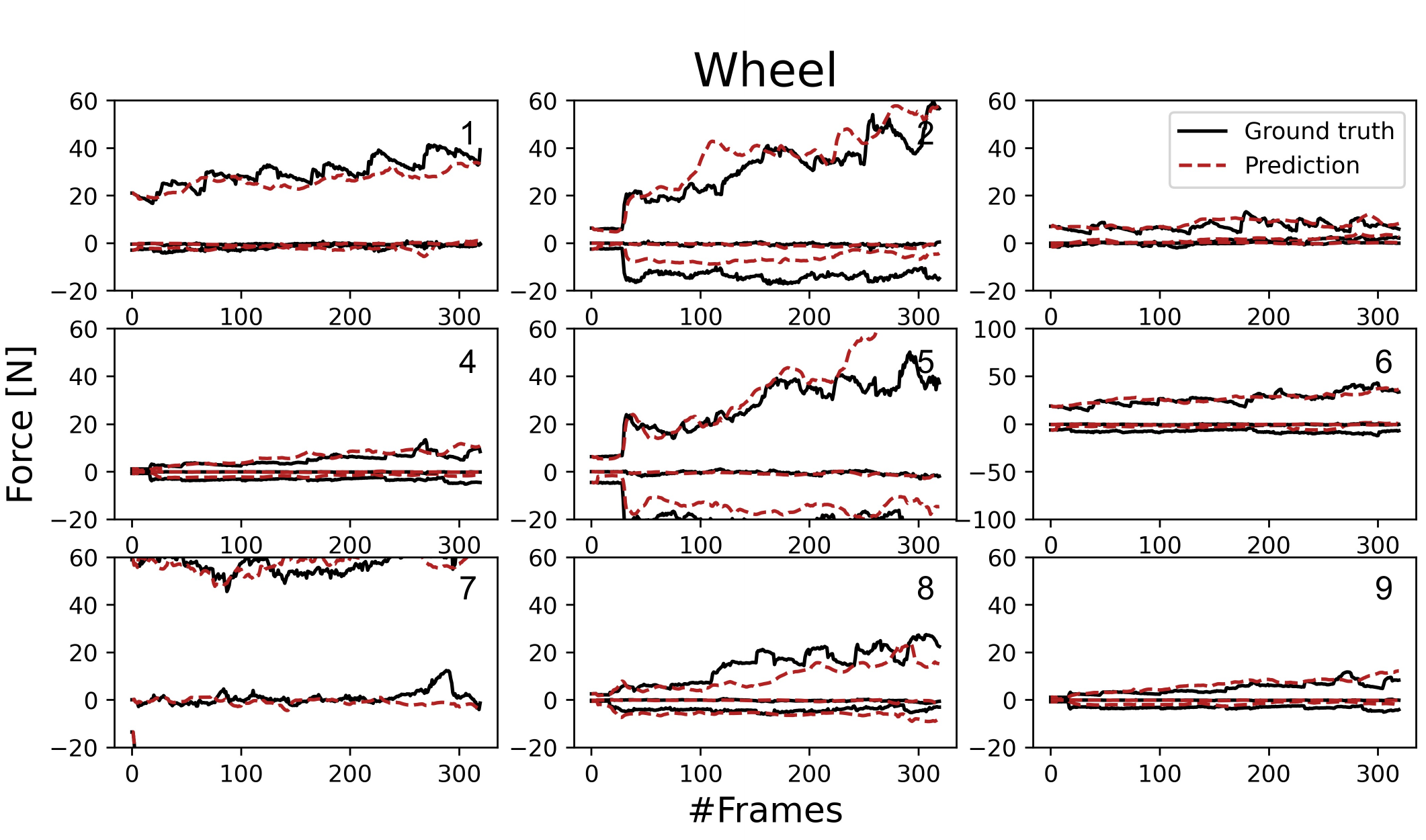}
        \caption{Interactions forces in 9 Wheel test examples. Each plot contains forces in forward, vertical and lateral directions.}
        \label{fig:forces_wheel_test}
    \end{figure}

    Finally, the time history of Excavation and Wheel interaction forces in 9 test examples are shown in Figures \ref{fig:forces_excav_test} and \ref{fig:forces_wheel_test}, respectively. Each plot of the figures include forces in forward, vertical and lateral directions. In most of the examples, the predictions are very well matched with the ground truths (in the test split).
    
\clearpage
    \setcounter{page}{23}

\end{document}